\documentclass{article}
\usepackage{microtype}
\usepackage{graphicx}
\usepackage{subcaption}
\usepackage{booktabs} 
\usepackage[bottom]{footmisc}
\usepackage{hyperref}

\usepackage[preprint]{icml2026}
\usepackage{amssymb}
\usepackage{mathtools}
\usepackage{amsthm}
\usepackage[table]{xcolor}   
\usepackage{listings}
\usepackage{pifont}
\usepackage{tcolorbox}
\tcbuselibrary{breakable}
\usepackage{makecell}

\usepackage[capitalize,noabbrev]{cleveref}

\theoremstyle{plain}

\theoremstyle{definition}

\theoremstyle{remark}

\usepackage[textsize=tiny]{todonotes}

\icmltitlerunning{MobiFlow: Real-World Mobile Agent Benchmarking through Trajectory Fusion}

\begin{document}

\twocolumn
[
  \icmltitle{\textit{MobiFlow}: Real-World Mobile Agent Benchmarking through Trajectory Fusion}




  \icmlsetsymbol{equal}{*}
  \begin{icmlauthorlist}
    \icmlauthor{Yunfei Feng}{yyy}
    \icmlauthor{Xi Zhao}{yyy}
    \icmlauthor{Cheng Zhang}{yyy}
    \icmlauthor{Dahu Feng}{xxx}
    \icmlauthor{Daolin Cheng}{yyy}
    \icmlauthor{Jianqi Yu}{zzz}
    \icmlauthor{Yubin Xia}{yyy}
    \icmlauthor{Erhu Feng}{yyy}
  \end{icmlauthorlist}
  \icmlaffiliation{yyy}{Institute of Parallel and Distributed Systems (IPADS), Shanghai Jiao Tong University}
  \icmlaffiliation{xxx}{Department of Precision Instrument, Tsinghua University}
  \icmlaffiliation{zzz}{ National Innovation Institute of High-end Smart Appliances}
  \icmlcorrespondingauthor{Erhu Feng}{fengerhu1@sjtu.edu.cn}
  \icmlkeywords{Machine Learning, ICML}
    {%
        \centering
        \captionsetup{type=figure}
        \includegraphics[width=1.0\textwidth]{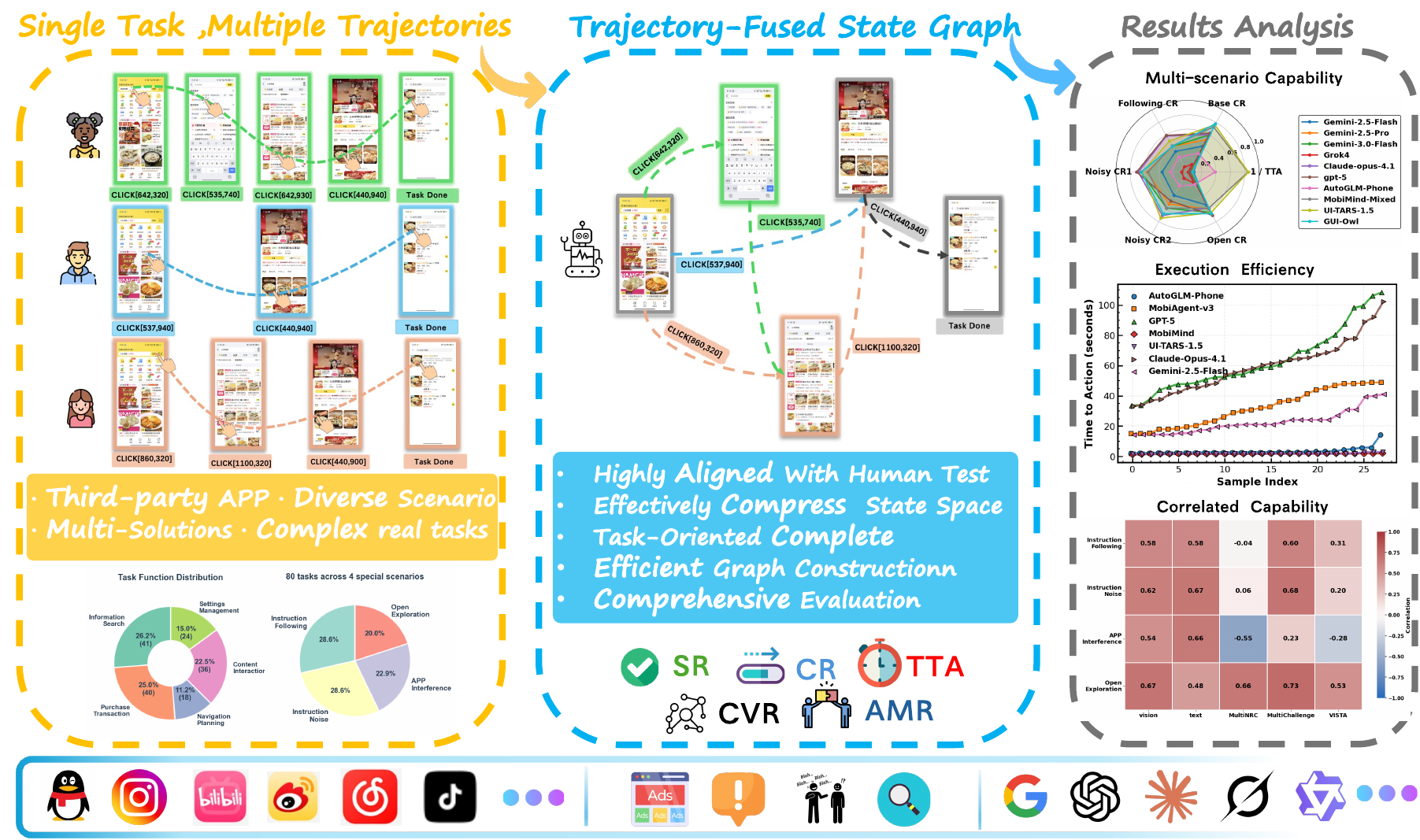}
        \captionof{figure}{
        \textbf{The framework of \textit{MobiFlow}.} It constructs state transition graphs from third-party applications and evaluates 20 applications across 240 tasks. The framework extends existing metrics to assess multiple models in terms of task completion, execution efficiency, generalization, and alignment with competency requirements. The code is available at \url{https://github.com/nanookfyf/MobiBench}. Our data will be released upon acceptance. 
        }
        \label{fig:overall}
    }
]

   \printAffiliationsAndNotice{}   

\begin{abstract}
Mobile agents can autonomously complete user-assigned tasks through GUI interactions. However, existing mainstream evaluation benchmarks, such as AndroidWorld, operate by connecting to a system-level Android emulator and provide evaluation signals based on the state of system resources. In real-world mobile-agent scenarios, however, many third-party applications do not expose system-level APIs to determine whether a task has succeeded, leading to a mismatch between benchmarks and real-world usage and making it difficult to evaluate model performance accurately. To address these issues, we propose \textit{MobiFlow}, an evaluation framework built on tasks drawn from arbitrary third-party applications. Using an efficient graph-construction algorithm based on multi-trajectory fusion, \textit{MobiFlow} can effectively compress the state space, support dynamic interaction, and better align with real-world third-party application scenarios. \textit{MobiFlow} covers 20 widely used third-party applications and comprises 240 diverse real-world tasks, with enriched evaluation metrics. Compared with AndroidWorld, \textit{MobiFlow}'s evaluation results show higher alignment with human assessments and can guide the training of future GUI-based models under real workloads. 
\end{abstract}

\vspace{-15pt}
\section{Introduction}
With the advancement of artificial intelligence technology, graphic User Interface (GUI) agents \citep{ye2025mobileagentv3fundamentalagentsgui}, driven by multimodal large models\citep{ma2024coco}, are emerging as a key technology poised to transform next-generation terminal applications. These agents, guided by human instructions, automatically accomplish daily and professional tasks across diverse device environments, thereby enhancing production efficiency and improving user operational experience. However, how to conduct fast and accurate evaluations of GUI agents' actual potential based on real-world usage scenarios remains a significant challenge\citep{rawles2024androidworld}.

Existing mobile-use evaluation benchmarks are primarily evaluated through online and offline methods. Online benchmark enables interaction with agents via Android emulators and extracts evaluation signals from system-level resource states\citep{toyama2021androidenv,rawles2024androidworld}. However, the vast majority of third-party applications do not expose system-level interfaces, making it difficult to obtain accurate evaluation signals and consequently limiting the range of applications that can be evaluated. For the Apps that can be evaluated, the environmental factors also make experimental reproduction difficult. Moreover, for vendor applications that do provide system-level APIs, mobile agents operating through GUI interactions are of limited practical relevance. As a result, such evaluation settings inevitably exhibit a gap from real-world usage scenarios.

Offline benchmarks rely on pre-collected human interaction trajectories and assess agents by comparing their action sequences against these reference trajectories\citep{zhang2024large,xu2025mobile}. While this approach can be applied to arbitrary applications, the state space of mobile user interfaces is nearly infinite, making it difficult for offline data collection to cover all possible states. Consequently, existing offline datasets typically provide only one or a few trajectories per task. Some studies\citep{song2025colorbench} attempt to construct a complete interface state transition graph by traversing all interactive UI elements; however, as the graph depth increases, the state space grows exponentially. Consequently, existing mobile usage evaluation benchmarks still struggle to assess agent performance on real-world third-party applications comprehensively.

\textit{To address the dilemma that online benchmarks fail to cover the full range of applications, while offline benchmarks fail to capture all task completion trajectories}. We observe that for each task, the set of independent UI objects relevant to the task on a single screen is finite (usually $<3$), and different trajectories often converge to the same node after only a few steps. Based on this insight, we propose \textit{MobiFlow}, which compresses the state space by constructing \textbf{Trajectory-Fused State Graphs}, enabling effective evaluation signals for arbitrary third-party applications while preserving complete state coverage.

\begin{figure}[ht]
    \centerline{\includegraphics[width=\columnwidth]{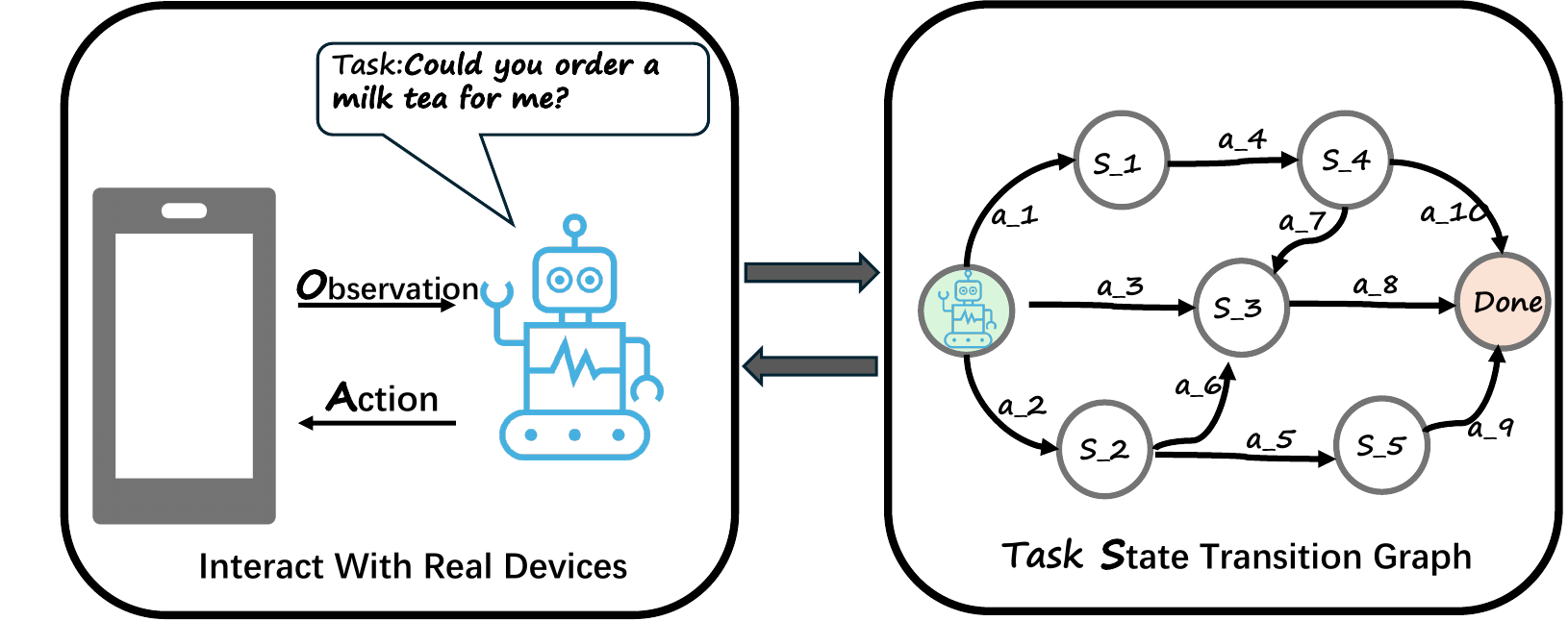}}
    \setlength{\belowcaptionskip}{-10pt}
    \caption{
      \textbf{Modeling agent-device interactions with state transition graph}. Executing actions triggers state transitions. Completing a task corresponds to reaching a terminal state.
    }
    \label{fig:graph}
\end{figure}

We efficiently collect human operation trajectories from real-world task scenarios using front-end tools\footnote{Implementation details are provided in the Appendix\ref{app:collect}.}, including interface screenshots, actions, and annotation information. By merging nodes with identical annotation information and sharing their action transitions, we consolidate multiple real trajectories into a state transition graph capable of simulating realistic tasks. This construction effectively models real-world interaction environments while compressing state complexity, enabling a comprehensive evaluation of mobile agents’ practical capabilities.

\textit{MobiFlow} covers 20 widely used mobile applications and includes 240 real-world tasks designed to comprehensively evaluate the performance of current mainstream general-purpose models and GUI-specialized models on practical tasks\citep{ye2025mobileagentv3fundamentalagentsgui,qin2025ui, wang2025ui,zhang2025mobiagent}. We abstract the agent’s task execution process as state transitions on a graph (see figure\ref{fig:graph}), which enables more precise evaluation\citep{bu2025limits}. We can establish new metrics, such as coverage rate and completion rate, to comprehensively measure model capabilities. In addition, we design specialized scenarios to assess specific model capabilities, including instruction-following, instruction-noise interference, and multi‑application interference, etc.

Through the aforementioned definitions and scenario design, we show that the metrics derived from the graph structure exhibit strong interpretability. On our benchmark, UI-TARS and GUI-OWL achieve success rates of 60.4\% and 55.7\%, respectively, with single-action execution times of 1.75 s and 20.02 s. Compared to other benchmarks, our evaluation results align more closely with human judgment. Moreover, \textit{MobiFlow} has been deployed to assess the capabilities of GUI agent models on smartphones with sales volumes exceeding ten million units.

\vspace{-5pt}
In summary, our main contributions are as follows:

\vspace{-10pt}
\begin{itemize}
    \item \textbf{An efficient multi-trajectory fusion-based graph construction algorithm}. Our algorithm effectively compresses the state space size, reduces complexity, and ensures generalization to real-world tasks.
    
    \vspace{-5pt}
    \item \textbf{Diverse evaluation metrics and test scenarios}. Our method evaluates multiple dimensions, including time efficiency and coverage, and designs specialized scenarios, such as instruction following and noise interference.

    \item \textbf{Effective guidance for the future development of mobile agents.} We conduct comparative evaluations between general-purpose models and specialized models on the evaluation set, along with an attribution analysis of their performance across different scenarios.
\end{itemize}
\vspace{-15pt}
\section{Related Work}
This section review existing evaluation frameworks(Detailed comparisons are provided in Table\ref{tab:datasets}), which can be broadly categorized into two types: offline evaluation based on pre-collected human interaction trajectories and online evaluation based on system-level emulators.

Current offline evaluation methods perform step-by-step evaluation by comparing the predicted actions against reference action trajectories. AITW\citep{rawles2023androidinthewild} introduced large-scale data for training and evaluation. AITZ\citep{zhang-etal-2024-android} refined AITW's data, resulting in a more concise dataset. ANDROIDCONTROL\citep{leung2025androidcontrol} and AMEX\citep{chai-etal-2025-amex} focused on single-step action accuracy via element or
coordinate matching. Mobile-Bench-v2\citep{xu2025mobile} covers a wide range of real-world tasks by collecting multiple possible trajectories and assessing them individually, which fails to support effective interaction. ColorBench\citep{song2025colorbench} constructs the complete state transition space of an application, leading to an explosion of the state space. Although existing offline evaluation methods\citep{lee2024benchmarking,chai2025a3} can cover arbitrary scenarios and conveniently provide valid evaluation signals, they still fail to effectively encompass diverse dynamic interactions.

Several researchers have proposed online evaluation systems that enable interaction with agents via Android emulators
and extract evaluation signals from system-level resources
state. Mobile‑Env\citep{zhang2024mobile} covers only 74 tasks, which limits its diversity. AndroidArena\citep{xing2024understanding} and AndroidWorld\citep{rawles2024androidworld} expand the range of tasks; they are confined to built‑in system applications whose designs often differ significantly from mainstream apps. AndroidLab \citep{xu2025androidlab} is similarly constrained by its reliance on specific applications. Other works, such as SPA‑Bench\citep{chen2024spa}, AndroidDaily\citep{lee2025benchmarkingmobiledevicecontrol} and MobileWorld\citep{kong2025mobileworld}, still suffer from several limitations, including unstable environments and highly limited evaluation signals. 

To address the dilemma that online benchmarks fail to cover the full range of applications, while offline benchmarks fail to capture all task completion trajectories, we aim to propose an evaluation framework that ensures coverage of diverse interactions while encompassing arbitrary applications and achieving high alignment with human evaluation.

\begin{table*}[h!]
\vspace{-2mm}
\centering
\resizebox{\textwidth}{!}{
\begin{tabular}{l c  c  c  c  c c c }
\toprule
Benchmark & \#APP & \#Task  & Interactive & Real App & No Dependencies  & Multi-Solution & Evaluation  \\
\midrule
Mobile-Bench-v2\citep{xu2025mobile} & 49 &12,856 & \textcolor{red}{\ding{55}} & \textcolor{green}{\ding{51}}  & \textcolor{green}{\ding{51}}  & \textcolor{green}{\ding{51}} & Trajectory-based \\
AndroidControl\citep{leung2025androidcontrol} & - &  14,548 & \textcolor{red}{\ding{55}} & \textcolor{green}{\ding{51}}  & \textcolor{green}{\ding{51}}  & \textcolor{red}{\ding{55}} & Trajectory-based \\
MobileAgentBench\citep{wang2024mobileagentbench} & 100 & 10 & \textcolor{green}{\ding{51}} & \textcolor{red}{\ding{55}} & \textcolor{red}{\ding{55}} & \textcolor{green}{\ding{51}} &Result-based \\
AndroidWorld\citep{rawles2024androidworld} & 20 & 116 & \textcolor{green}{\ding{51}} & \textcolor{red}{\ding{55}} & \textcolor{red}{\ding{55}} & \textcolor{green}{\ding{51}} &Result-based   \\
AndroidLab\citep{xu2025mobile}   &  9 & 138 & \textcolor{green}{\ding{51}} & \textcolor{red}{\ding{55}} & \textcolor{red}{\ding{55}} & \textcolor{green}{\ding{51}} &Result-based \\
SPA-Bench \citep{chen2024spa}&  66  & 340 & \textcolor{green}{\ding{51}} & \textcolor{green}{\ding{51}} & \textcolor{red}{\ding{55}} & \textcolor{green}{\ding{51}} &Result-based \\
\midrule
\textit{\textit{MobiFlow}} & 20 & 240 & \textcolor{green}{\ding{51}} & \textcolor{green}{\ding{51}} & \textcolor{green}{\ding{51}} & \textcolor{green}{\ding{51}} & Graph-based   \\
\midrule
\bottomrule

\end{tabular}

}
\caption{ \textbf{Comparison of different datasets and environments
for benchmarking Mobile GUI agents.} Column definitions: \#Task. (number of tasks), \#Apps (number of applications),
Interactive (Support environmental interaction), Real App (Including any third-party apps), No Dependencies (Does not need additional software dependencies, such as an Android emulator), Multi-Solution (supports solving tasks in multiple ways), Evaluation (Evaluation
Strategy) }  
\label{tab:datasets}
\vspace{-0.3in}
\end{table*}

\section{Formulation}
We provide a formal description of how a Mobile agent accomplishes a given task.
Let $\mathbf{G}$ denote the set of all tasks, and consider a specific task
$\mathbf{g} \in \mathbf{G}$. We model the mobile device together with the
application software as an observable finite-state machine:
\begin{equation}
\mathbf{\mathcal{M}_g} = (\mathbf{\mathcal{S}}, \mathbf{\mathcal{A}},
\mathbf{\mathcal{O}}, \mathbf{\mathcal{T}}, \mathbf{\mathcal{R}}),
\label{eq:mdp_gui}
\end{equation}
where $\mathcal{S}$ is a finite set of states and $\mathcal{A}$ is a finite set
of actions\footnote{Details of the action space are provided in
Appendix~\ref{app:act_space}.}. Each action $a \in \mathcal{A}$ corresponds to a
basic UI operation, such as clicking, swiping, or text input. $\mathcal{O}$
denotes the observation space, where the observation $o \in \mathcal{O}$
perceived by the agent is determined by the current state $s \in \mathcal{S}$.
The state transition function $\mathcal{T}: \mathcal{S} \times \mathcal{A}
\rightarrow \mathcal{S}$ is assumed to be deterministic(the UI transition structure of APPs remains fixed, even though its content may vary randomly), and
$\mathcal{R}$ represents the environment reward function.
\begin{itemize}
    \item A statistical analysis that motivates the task-oriented environment
    formulation is presented later in this paper.
    \item The transition function $\mathcal{T}$ induces a directed graph
    $\mathcal{G} = (\mathcal{V}, \mathcal{E})$, where an edge
    $e = (s, s') \in \mathcal{E}$ exists if and only if there is an action
    $a \in \mathcal{A}$ such that $\mathcal{T}(s, a) = s'$. Each edge is labeled
    with the corresponding action $a$.
\end{itemize}

We define a Mobile agent as a decision-making entity equipped with internal
reasoning and memory mechanisms:
\begin{equation}
\mathbf{\mathcal{AG}} = (\mathbf{\Sigma}, \mathbf{\Pi}, \mathbf{\mathcal{H}}),
\label{eq:agent_model}
\end{equation}
where $\mathbf{\Sigma}$ denotes the reasoning space, updated according to
$\sigma_t \sim \mathbf{\Sigma}(\cdot \mid h_{t-1}, o_t)$; $\mathbf{\Pi}$ denotes
the decision space, from which actions are sampled via
$a_t \sim \mathbf{\Pi}(\cdot \mid h_{t-1}, \sigma_t, o_t)$; and
$\mathbf{\mathcal{H}}$ denotes the memory space, which is updated as
$h_t = h_{t-1} \cup \{o_t, \sigma_t, a_t\}$.

\begin{itemize}
    \item In practice, the reasoning and decision spaces are often integrated
    into a single module in many models; we present them separately here for
    conceptual clarity.
    \item Depending on the implementation, the memory update process may not
    explicitly store all components—observation, reasoning state, and action.
\end{itemize}


\section{MobiFlow}
To capture state transitions across diverse tasks, accurately evaluate agent performance under varied scenarios, and enable in-depth attribution analysis of agent capabilities, we propose the \textit{MobiFlow} framework. This section is organized into three parts, detailing our innovations in evaluation environment construction, metric design, and specialized scenario development, respectively.

\subsection{Task Graph Construction}

\begin{figure}[h]
    \centerline{\includegraphics[width=\columnwidth]{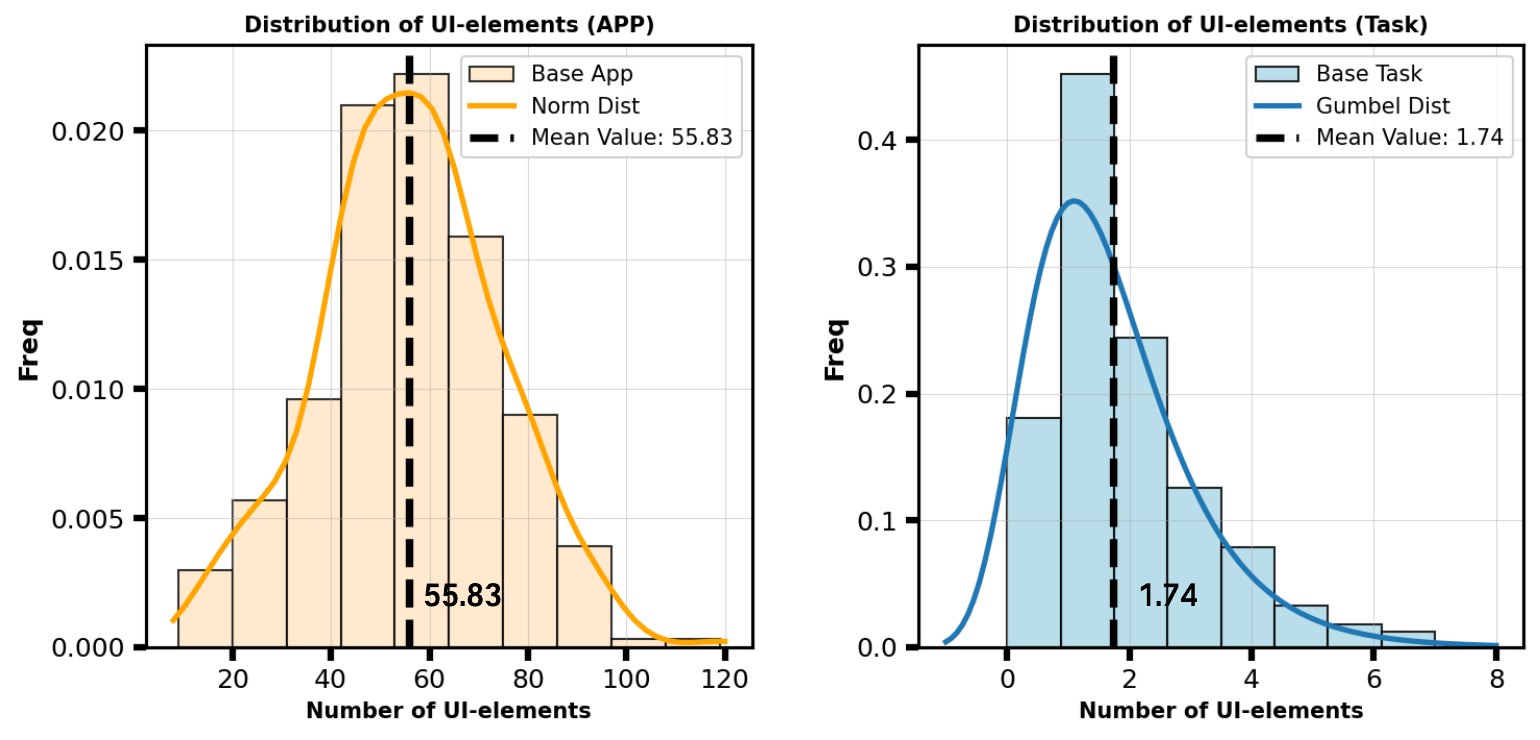}}
    \caption{
      \textbf{Comparison of the Distribution of Interactive Element Counts Across Different Modeling Approaches}. The left figure presents the app-based modeling approach, while the right figure illustrates the task-based modeling approach.
    }
    \label{fig:act_num}
    \vspace{-2mm}
\end{figure}

\textbf{Observations}. Constructing graphs via exhaustive search leads to state space explosion, while directly building complete graphs incurs extremely high construction complexity and human labor costs. Through an analysis of the top 30 mainstream apps\footnote{Static details in Appendix\ref{app:statistic}}, we find that the number of independent interactive elements on an app interface approximately follows a normal distribution with a mean of 55.8(see fig\ref{fig:act_num}). In contrast, for a single task, the number of task-relevant elements approximately follows a Gumbel extreme value distribution, with a mean of only 1.7. This indicates that task-oriented graph construction yields a state space complexity($O(1.7^d)$, where $d$ is action depth) that is far smaller than that of search-based approaches($O(55.8^d)$). Moreover, we observe that most trajectories converge to states with similar transition structures after only a few steps.
\vspace{-5pt}

\vspace{-10pt}
\textbf{Correctness and Completeness}. Considering the requirements of industrial deployment and the characteristics of real-world applications, we aim to construct a \textbf{Task-Oriented Complete State-Transition Graph (TCSG)}, which includes all correct, task-relevant state transitions while excluding irrelevant and redundant states. In real-world deployments, task success rates are typically required to exceed 95\%, and triggering incorrect, task-irrelevant transitions is undesirable, as additional action paths can reduce both execution efficiency and success rates. Accordingly, we focus on whether an agent can sample correct actions to complete the task. For incorrect actions, we handle them by keeping the interface unchanged, transitioning to a blank screen, or presenting task-specific UI prompt interfaces.
 
\begin{figure}[h]
    \centerline{\includegraphics[width=\columnwidth]{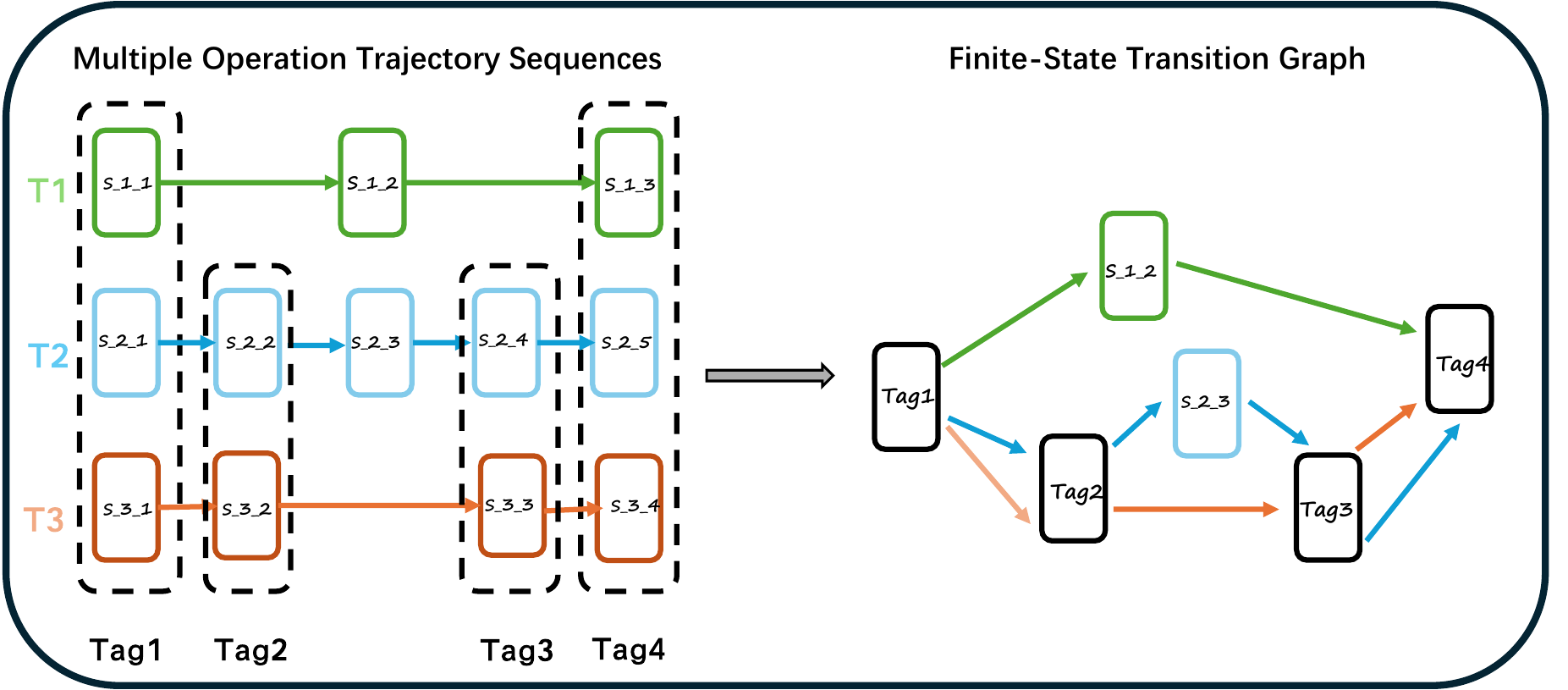}}
    \caption{
      \textbf{Algorithm workflow illustration:} For states with identical labels (or equivalently, similar transition structures), we merge them and allow the merged state to share their transition conditions, thereby compressing the state complexity.
    }
    \label{fig:build_graph}
\end{figure}

\textbf{Trajectory Fusion.} We observe that different trajectories tend to converge to states with highly similar transition structures after very few steps, indicating the presence of many reusable state transitions. This reuse can further reduce the complexity of the state space. For a single task, we collect all directly completed trajectories and assign consistent labels to states that share identical transition structures across these trajectories. States with the same labels are then merged to share transition relations, thereby enabling low‑complexity TCSG construction (as shown in Figure \ref{fig:build_graph}). For cross-application tasks, we can likewise construct via trajectory merging or multi-graph connectivity methods. 

\vspace{-5pt}

\vspace{-10pt}
\textbf{Efficient Graph Construction}.
The construction of the TCSG can be mainly divided into three stages. First, trajectories are traversed to assign labels to states. Second, chain-structured transitions are initialized. Finally, the transition spaces of states sharing the same labels are merged via union coverage\footnote{The construction algorithm for the transition graph is provided in Appendix\ref{app:graph_build}}. We achieve fast localization of target UI elements by parsing XML contents. More efficient labeling is enabled through UI structural similarity matching and visual model inference. We employ efficient node representations and maintain a union-find structure to support fast merging of transition spaces. With the techniques described above, the time cost of practical data collection and graph construction can be substantially reduced. For example, a task that involves posting a comment on Weibo typically requires around 15 interaction steps to complete. By collecting 7 trajectories, we can cover all valid task completion paths, and the entire construction process, including manual verification, can be finished in under 10 minutes.

\vspace{-10pt}
\subsection{Comprehensive Evaluation}

Based on the aforementioned construction algorithm, a TCSG encompassing all reasonable behavioral trajectories can be rapidly generated for a given task. As the graph is constructed from a global perspective, it enables a more comprehensive evaluation of model behavior. In contrast, most previous evaluation methods, constrained by paradigms that rely solely on outcome signals or single-trajectory analysis, primarily focus on task success rate (SR)\citep{xie2024osworld}. This metric considers only task outcomes and ignores the execution process, which limits its ability to fully assess the model’s capabilities.

To more accurately evaluate task completion on the TCSG, we introduce four complementary metrics. Completion Rate (CR) measures the agent’s progress toward the goal state, while Coverage Rate (CVR) quantifies the proportion of explored states, reflecting the agent’s exploratory capability. Action Match Rate (AMR) assesses the alignment between the agent’s executed actions and the instructed actions, serving as an indicator of instruction compliance and safety. Time to Action (TTA) evaluates execution efficiency. Together, these metrics form a comprehensive evaluation framework spanning four dimensions: task progression, state exploration, instruction compliance, and execution efficiency. A detailed comparison and summary of these metrics is provided in Table\ref{tab:metrics}.

\vspace{-10pt}
\subsection{Diverse Scenario}
\begin{figure}[H]
    \vspace{-2mm}
    \centerline{\includegraphics[width=\columnwidth]{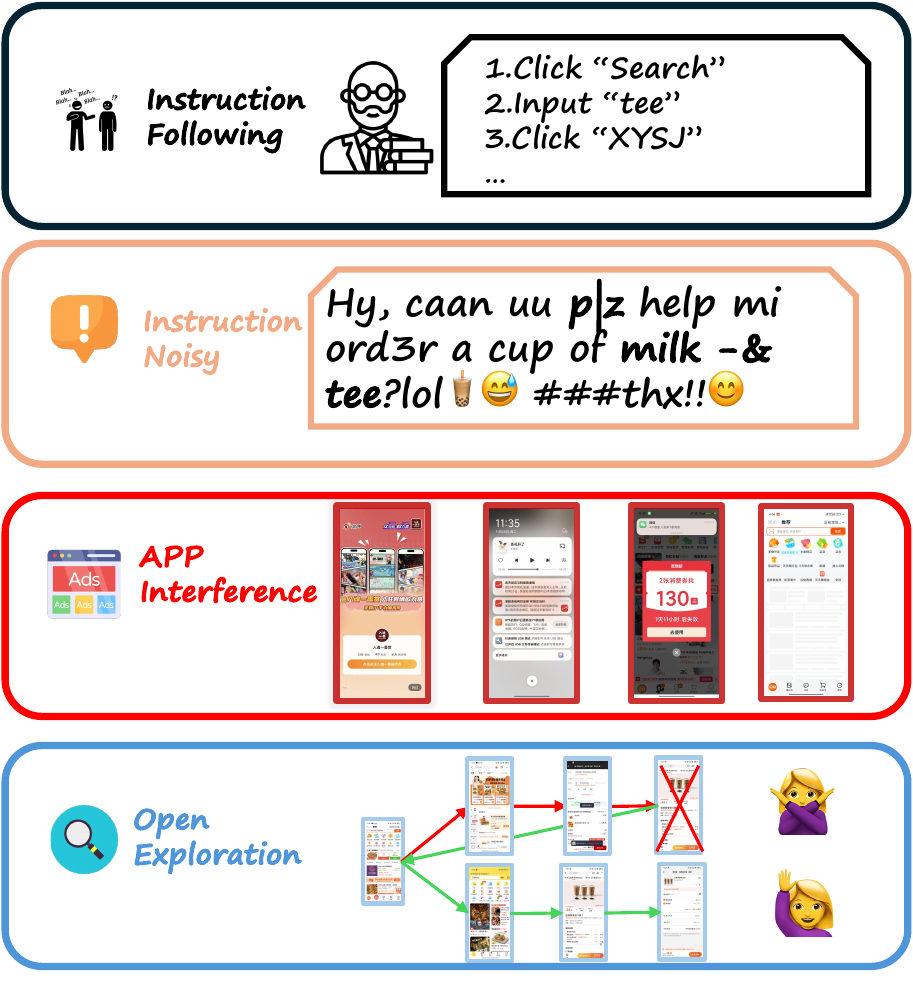}}
    \setlength{\belowcaptionskip}{-15pt} 
    \caption{
      \textbf{Special scenarios} including instruction following, instruction interference, application interference, and open exploration (to observe whether the model can deviate from erroneous paths).
    }
    \label{fig:scenarios}
\end{figure}

In this section, we introduce a variety of evaluation scenarios constructed based on task-state transition graphs.(see fig \ref{fig:scenarios}) These scenarios are designed to assess model performance from multiple perspectives, including practicality, safety, robustness, and exploratory capability.

\textbf{Base Scenario}: Based on construction algorithms, we constructed 160 distinct tasks spanning 20 mainstream applications. To ensure these tasks reflect real-world usage scenarios, we selected tasks across multiple domains, including common activities like information search, product purchasing, ride-hailing navigation, social chatting, and application settings. Furthermore, we resampled task descriptions to ensure comprehensive alignment with real-world applications. This framework enables a thorough evaluation of agents' capabilities for handling common everyday tasks.

\textbf{Special Scenario 1 Instruction Following}: The agent should be able to understand and adhere to human instructions effectively. In this scenario, specific execution steps will be provided in the instructions. The action matching rate(AMR) will be used to evaluate the agent's ability to follow instructions and its understanding and memory of complex, lengthy instructions. Additionally, this scenario can also assess the agent’s safety performance.

\textbf{Special Scenario 2 Instruction Noise Interference}: In daily communication, errors such as typos or the use of emojis and special characters may occur. The agent should be able to identify the true task intent within such noisy instructions and correctly complete the task. In this scenario, noise will be introduced into task instructions through methods such as common typo substitutions, emoji injections, and multilingual mixing. This aims to evaluate the model's resistance to interference and the robustness of its instruction comprehension.

\begin{table*}[t]
\vspace{-2mm}
\footnotesize
\setlength{\tabcolsep}{4pt}
\renewcommand{\arraystretch}{1.15}
\begin{tabular}{p{1.8cm} p{2.6cm} p{5.8cm} p{5.8cm}}
\toprule
\textbf{Metric} 
& \textbf{Formula} 
& \textbf{Notation Explanation} 
& \textbf{ Description} \\
\midrule

Success Rate (SR)
&
$\frac{1}{N}\sum_{i=1}^{N}\text{suc}(i)$
&
$N$: number of tasks; $\text{suc}(i)=1$ if $g_i$ is completed, else $0$. 
&
Indicates whether a task is fully completed, focusing only on the final outcome.
\\

\midrule

Completion Rate (CR)
&
$\max\!\left\{
\frac{d(s_{\text{visited}}, s_{\text{goal}})}
{d(s_{\text{start}}, s_{\text{goal}})}
\right\}$
&
$s_{\text{start}}, s_{\text{goal}}$: initial and goal states;
$d(\cdot,\cdot)$: shortest-path distance.
&
Measures the maximum progress achieved toward the goal state.
\\

\midrule

Coverage Rate (CVR)
&
$\frac{\sum_i |\mathcal{S}_{\text{visited}}|_i}
{\sum_i |\mathcal{S}|_i}$
&
$|\mathcal{S}_{\text{visited}}|_i$: visited states in task $i$;
$|\mathcal{S}|_i$: total states. 
&
Evaluates exploration by the proportion of states observed.
\\

\midrule

Action Match Rate(AMR)
&
$\frac{1}{N}\sum_i
\frac{\sum_j am[i][j]}{|IF_i|}$
&
$|IF_i|$: number of instructions;
$am[i][j]=1$ if instruction $j$ is satisfied. 
&
Measures instruction-following ability and safety.
\\

\midrule

Time to Action (TTA)
&
$\frac{1}{N}\sum_i
\frac{\sum_j t_{ij}}{|\tau_i|}$
&
$|\tau_i|$: number of interactions;
$t_{ij}$: time to generate action $j$. 
&
Captures execution efficiency via average action latency.
\\

\bottomrule
\end{tabular}
\caption{\textbf{Evaluation metrics defined on the state-transition graph.}}
\label{tab:metrics}
\end{table*}
\vspace{-5pt}
\begin{figure*}[h!]
    \centerline{\includegraphics[width=\textwidth]{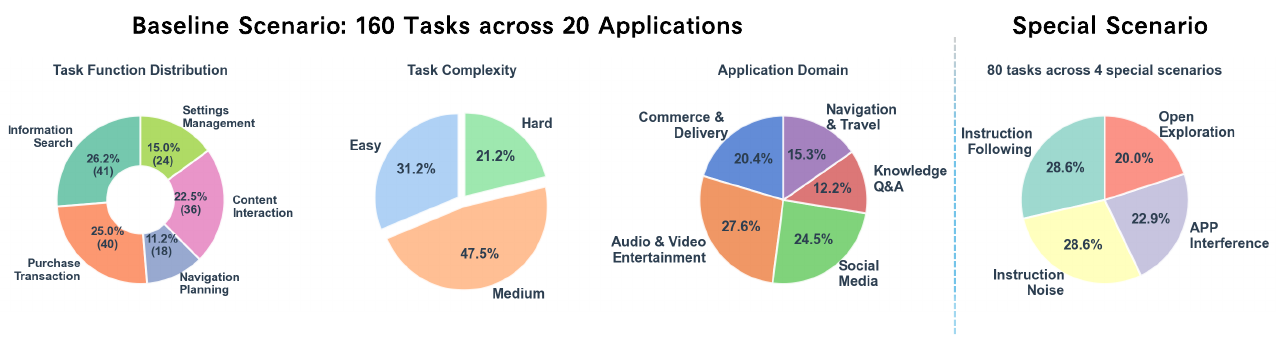}}
    \setlength{\belowcaptionskip}{-15pt}
    \setlength{\abovecaptionskip}{-1pt}
    \caption{
       \textbf{Task statistics for \textit{MobiFlow}}, covering the type, complexity, and application domain of basic-scenario tasks, as well as the quantity of special-scenario tasks.
    }
    \label{fig:taskdata}
\vspace{-2pt}
\end{figure*}

\textbf{Special Scenario 3 APP Interference}: In real-world applications, software often faces various interferences, such as pop-up notifications from other apps, ad pushes, unintended touches, or interface jumps caused by system anomalies. The agent should respond safely when encountering such abnormal situations. Therefore, instances containing these interferences are collected separately to evaluate the model's ability to cope with specific APP disturbances.

\textbf{Special Scenario 4 Open Exploration}: A knowledgeable agent should possess exploration and autonomous learning capabilities. To this end, we artificially constructed a set of special transition graphs. For example, based on basic scenarios, we introduce erroneous paths that do not satisfy task requirements, allowing us to observe whether the model can detect issues within these incorrect paths and reasonably backtrack to accomplish the designated task.

\begin{table*}[h]
\vspace{-2mm}
\caption{\textbf{Main Results on \textit{MobiFlow}}.Includes general-purpose and specialized models. TTA (Time to Action), SR (Success Rate), CR (Completion Rate), AMR (Action Matching Rate), CVR (Coverage Rate). CR1 (with instruction noise), CR2 (under APP interference).}
\resizebox{\textwidth}{!}{

\begin{tabular}{lcc|ccc|cc|cc|cc}
 \toprule
 \multicolumn{3}{c}{} & \multicolumn{3}{c}{Base Scenario} & \multicolumn{2}{c}{Following} & \multicolumn{2}{c}{Noisy Scenario} & \multicolumn{2}{c}{ Open Exploration }\\
  Model & Size(B) & TTA(s)$\downarrow$ & SR(\%)$\uparrow$ & CR(\%)$\uparrow$  & CVR(\%)   & AMR(\%)$\uparrow$& SR(\%)$\uparrow$  & CR1(\%)$\uparrow$ & CR2(\%)$\uparrow$  & CR(\%)$\uparrow$ & CVR(\%) \\ 
  \midrule
  \rowcolor{orange!10}
  \multicolumn{12}{c}{\textit{General Model}} \\
  Gemini-2.5-Flash & - & \underline{17.03} & 30.7 & 41.7 & 17.4  & 35.5 & 33.3 & 35.5 & 37.7  & 53.2 & 21.1 \\
  Gemini-2.5-Pro   & - & 30.12 & 38.7 & 52.1 & 26.9   & 55.3 & 41.6 & 56.4 & 51.7  & 59.7 & 28.1 \\
  Gemini-3.0-Flash & - & 33.31 & 40.2 & 59.7 & 30.2 & 56.4 & 41.6 & 64.7 & 58.4  & 65   & 34.3 \\ 
  Grok4            & - & 142.21 & 8.3 & 12.3 & 8.41  & 4.1 & 8.3 & 8.1 & 10.0  & 17.8   & 10.2 \\ 
  Claude-opus-4.1  & - & 46.68 & 43.6 & 60.3 & 34.6 & 62.2 & 58.3 & 68.6 & \underline{70.1}  & 57.9 & 37.7\\ 
  GPT-5            & - & 55.49 & \underline{49.7} & \underline{62.7} & 35.3  & \textbf{\underline{71.4}} & \textbf{\underline{59.3}} & \textbf{\underline{71.1}} & 46.7  & \textbf{\underline{70.0}} & 39.3 \\ 
  \rowcolor{blue!10}
  \multicolumn{12}{c}{\textit{GUI Model}} \\
  AutoGLM-Phone & 9B & 3.85 & 16.7 & 30.2  & 10.6   & 30.5 & 24.2 & 24.2 & 22.5 & 20.1 & 15.1 \\ 
  MobiMind-Mixed & 4B & 1.76 & 53.8 & 72.9  & 31.9  & 59.9 & 42.3 & 55.1 & 58.3  &  58.6  & 35.1 \\
  UI-TARS-1.5  & 8B & \textbf{\underline{1.75}} & \textbf{\underline{60.4}} & 76.3  & 34.4  & 63.1 & \underline{50.0} & 63.1 & \textbf{\underline{74.5}} & 65.7 & 37.1 \\
  GUI-Owl      & 8B & 20.02 & 55.7 & \textbf{\underline{77.9}}  & 37.5   & \underline{65.4} & 41.6 & \underline{64.7} & 67.4 & \underline{67.1} & 38.1 \\ 

\end{tabular}
}
\label{tab:main_results}
\vspace{-4mm}
\end{table*}
\vspace{-2mm}
\section{Experiments}
This section analyzes the evaluation results of several mainstream general and specialized models on our evaluation benchmark(see Table\ref{tab:main_results}). 
\vspace{-5pt}
\subsection{Experimental Setup}
\label{sec:set}
We evaluated a series of advanced models, including general-purpose models and GUI-specific models: OpenAI GPT-5, Claude-opus-4.1, Grok4, Gemini-2.5-flash, Gemini-2.5-Pro, Gemini-3-flash\citep{openai2025conversation,Anthropic2025Claude4.1,comanici2025gemini}, UI-TARS-1.5\citep{qin2025ui}, AutoGLM-Phone\citep{liu2024autoglm}, GUI-Owl\citep{ye2025mobile}, and MobiMind\citep{zhang2025mobiagent}. To ensure a consistent comparison, we adopted the same execution framework for general models, incorporating OminiParser\citep{lu2024omniparserpurevisionbased} to annotate actionable UI elements to reduce the complexity of coordinate-based interface operations\footnote{Detailed implementation can be found in the Appendix\ref{app:model_exec}}. For GUI-specific agents, we employed their official execution frameworks. The maximum number of interaction steps allowed was set to 50. Additionally, the screen resolutions of the evaluation data were primarily 1080×2400 and 1200×2670.

For specialized models, we standardized deployment and resource allocation, conducting inference deployment on NVIDIA A100-SXM4-80GB GPUs using the vLLM\citep{kwon2023efficient} inference framework. For general models, we uniformly utilized the OpenRouter API for requests, with the temperature parameter consistently set to 0.
\vspace{-5pt}
\subsection{Main Results}

\textbf{Task Completion}. In the base scenario and the APP interference scenario, GUI agents achieve superior performance. Specifically, GUI-OWL attains the highest completion rate (CR) of 60.4 in the base scenario, while UI-TARS achieves the highest CR of 74.5 under APP interference. However, specialized GUI models do not exhibit a comprehensive advantage over general-purpose models. In instruction-following, instruction-noise interference, and open exploration scenarios, general-purpose models (e.g., OpenAI GPT-5) demonstrate stronger capabilities.

\begin{figure}[h]
    \centerline{\includegraphics[width=\columnwidth]{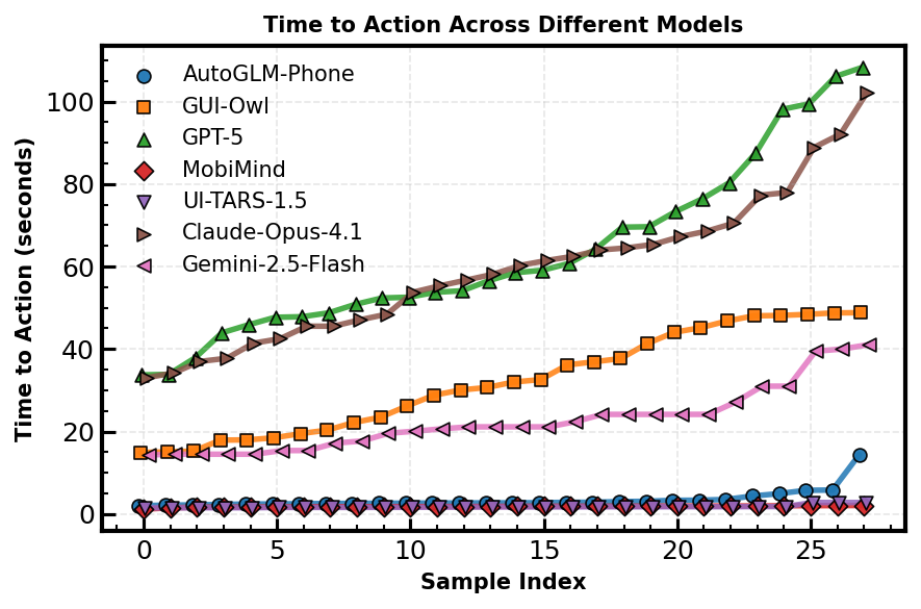}}
    \setlength{\belowcaptionskip}{-12pt}
    \caption{
      \textbf{Model execution efficiency.} GUI models use the same inference framework and operate under identical network request environments.
    }
    \label{fig:tta}
\vspace{-2mm}
\end{figure}

\textbf{Execution Efficiency}. We sampled tasks of varying complexity and analyzed the execution efficiency of agent interactions, as shown in Figure \ref{fig:tta}. The factors that affect single-action latency include network latency, model size, the number of decoding tokens, and the number of model invocations per action. Under the same model size, fewer decoding tokens generally lead to higher efficiency. For some models, such as GUI-OWL, a single action requires multiple model calls, which degrades action execution efficiency. In contrast, MobiMind and UI-TARS achieve efficiency advantages due to their lightweight architectures, single-call execution per action, and limited decoding outputs. General models, on the other hand, suffer from higher action latency because of greater network latency, larger model sizes, and decoding outputs that vary substantially with task difficulty\footnote{All experiments followed the settings in the section\ref{sec:set}.}.

\textbf{Performance Fluctuations}. We observed that the evaluation results of general models exhibit greater fluctuation than those of GUI models. Furthermore, while the execution paths of GUI models remain relatively consistent across multiple sampling attempts for a given task, general models demonstrate diverse paths but unstable completion outcomes, characterized by substantial variance.\footnote{Experiment details are provided in the Appendix\ref{app:tree}.}

\textbf{Trajectory Coverage}. Across multiple model evaluations, coverage and completion rates consistently show a positive correlation. Models with higher coverage also demonstrate better performance in open exploration scenarios.

\textbf{Resolution Scaling}. We analyze the impact of input image resolution on model performance\footnote{Details are provided in the Appendix\ref{app:rf}}. For images with a resolution of 1080×2400, the model’s task completion rate drops significantly when the resolution is scaled below 0.2×.

\subsection{In-Depth Analysis}
We observe that general models not only exhibit greater robustness against instruction noise interference compared to specialized GUI models, but also achieve a higher completion rate in instruction-following capability under such conditions. As shown in the table \ref {tab:instnoisy}, one possible explanation is that the presence of specific interference may heighten the model’s attention to the task-relevant portion of the long-context instructions. 

\begin{table}[h]
\small
\begin{tabular}{l |   c  | c }
\toprule
 Model &  \begin{tabular}[c]{@{}c@{}}No Instruction Noise\\ AMR \end{tabular}  & \begin{tabular}[c]{@{}c@{}} Instruction Noise \\ AMR \end{tabular} \\
\toprule
\multicolumn{3}{c}{\textbf{General Model}} \\
\midrule
Gemini-2.5-Flash & 35.55 & 38.55\scriptsize\color{red}{+3.0}  \\
Gemini-2.5-Pro   & 55.34 & 56.48\scriptsize\color{red}{+1.1}  \\
Claude-sonnet-4.5& 66.38 &67.38\scriptsize\color{red}{+1.0}  \\
GPT-5            &71.38  &71.07\scriptsize\color{green}{-0.3}  \\
\toprule
\multicolumn{3}{c}{\textbf{GUI specialized Model}} \\
\midrule
MobiMind-Mixed   & 59.89 & 55.11\scriptsize\color{green}{-4.7}  \\
UI-TARS-1.5 & 63.12 & 63.12\scriptsize\color{gray}{-0.0}   \\
GUI-OWL  & 65.37 & 64.72\scriptsize\color{green}{-0.4}   \\
\end{tabular}
\caption{\textbf{Robustness to Instruction Noise}: A Comparison of General Models vs. GUI Models}
\label{tab:instnoisy}
\vspace{-1mm}
\end{table}

General models exhibit strong generalization capabilities and outperform GUI models in exploratory tasks. However, lightweight specialized models support on-device deployment, while their stable accuracy in completing a large volume of tasks further facilitates practical deployment and application in real-world scenarios.

\begin{figure}[ht]
    \centerline{\includegraphics[width=\columnwidth]{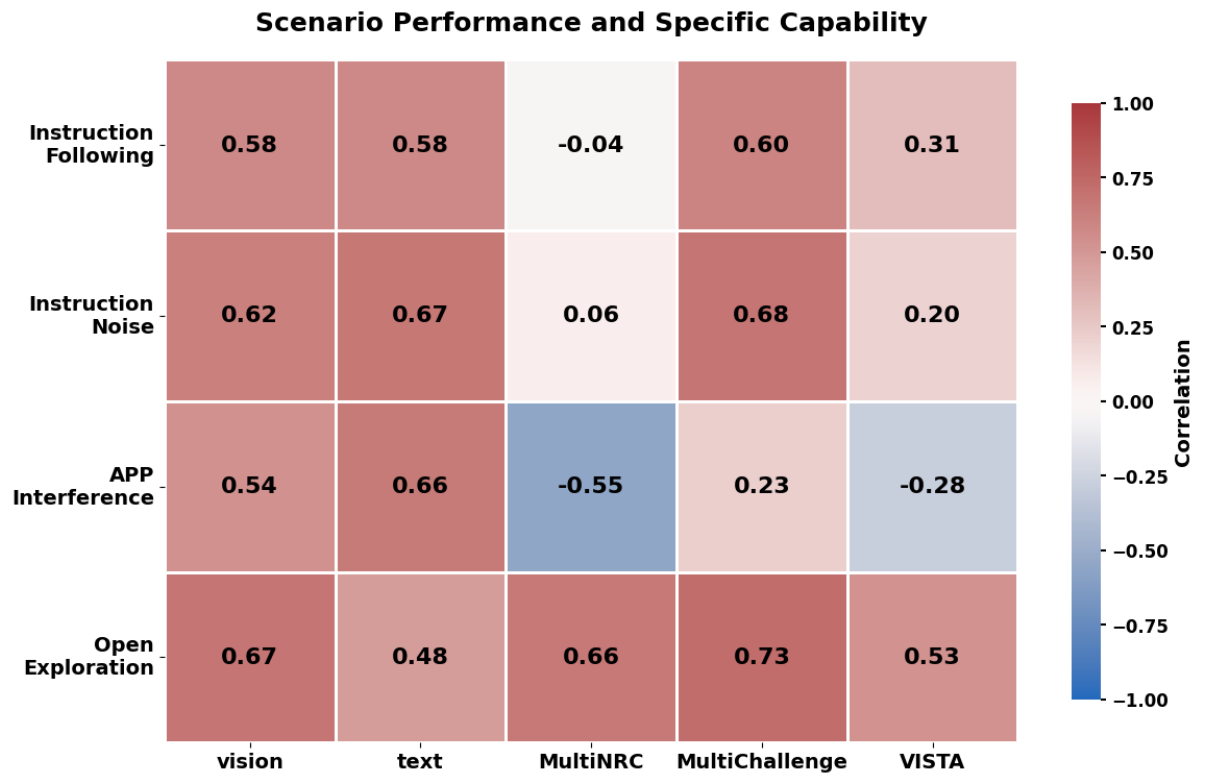}}
    \caption{
      \textbf{Correlation Analysis}. Correlation between General Models' Performance in Specific Scenarios and Different Benchmark Capabilities.
    }
    \label{fig:heat}
\end{figure}

Across different scenarios, we analyze the performance scores of general models and compare them with established benchmarks(Text\&VisionArena\footnote{\url{https://lmarena.ai}}, MultiNRC\citep{fabbri2025multinrc}, MultiChallenge\citep{he2024multi}, and VISTA\citep{zheng2023judging}\footnote{\url{https://scale.com/leaderboard}}) to investigate which specific capabilities are reflected by each scenario. Our correlation analysis results are illustrated in the figure \ref{fig:heat}.

We find that each scenario exhibits relatively high correlations with capabilities in visual input understanding and cross-cultural contextual comprehension. In open exploration and instruction-following scenarios, the correlation with multi-turn dialogue capability is particularly pronounced. Under app interference conditions, a certain negative correlation with logical reasoning capability is observed. Additionally, the open exploration scenario shows a particularly strong correlation with causal reasoning ability.

\subsection{Special issues}
\textbf{Thinking and observation become desynchronized}. During the experiments, we observed that GUI models occasionally perform erroneous actions without triggering the corresponding page transition. Yet, the models proceed under the assumption that the transition has been completed, continuing to operate as if they were in the post-transition state.

\textbf{Insufficient Generalization of Reasoning}. We observe that some models tend to operate based on fixed reasoning patterns. For instance, when encountering unconventional tasks, most models default to searching for task-related keywords, occasionally overlooking the information inherently available on the page itself.

\textbf{Interface Comprehension Deviation}. We have observed that models can sometimes be misled by interface information. For instance, in search scenarios, actual application recommendation algorithms often display suggested content in light colors; however, the model frequently misinterprets this as user-input text and attempts to clear it. 

\subsection{Trade-offs}
 \textbf{Balancing Agent Execution and System Boundaries.} Granting the agent greater autonomy over certain system operations can expand its action space and enhance overall capability. For instance, combining clicking and text-input actions can prevent the issue of repeated input field activation in search scenarios. However, such delegation must be carefully managed to ensure system security and stability.

 \textbf{Balancing Generalization and Reliability.} Strengthening an agent’s generalization ability leads to more diverse execution trajectories, which enriches behavioral flexibility. Yet, this often comes at the cost of reduced determinism and reliability in task completion. In real-world deployment, a deliberate and nuanced balance must be struck to maintain both adaptability and consistent performance.

\vspace{-10pt}
\section{Discussion}
\textbf{Cross-APP Tasks}. We support cross-application tasks via 
trajectory merging or the multi-graph connectivity method. In this paper, we focus on evaluating the intrinsic capabilities of GUI models, and therefore, only single-task results are reported.

\textbf{Reproduction}. Our state graph is configured with deterministic transitions to ensure reproducible and accurate evaluation of third-party applications.

\textbf{MCP Support}. Since this paper focuses more on the evaluation of GUI operations for agents, support for the Model Context Protocol (MCP)\citep{yan2025mcpworld} will be addressed in future work.

\section{Conclusion}
In summary, we introduce \textit{MobiFlow}, a mobile-use benchmark built on arbitrary third-party APPs. By constructing state transition graphs, \textit{MobiFlow} effectively compresses the state space of multiple task-completion trajectories. Compared with existing benchmarks, \textit{MobiFlow} more faithfully reflects the practical capabilities of agents. We equip the benchmark with a diverse set of evaluation metrics and specialized evaluation scenarios to enable in-depth analysis of the model's specific capabilities. We analyze the performance of current mainstream models and identify the challenges they face. Our benchmark provides a critical solution for gaining a deeper understanding of the real-world capabilities of mobile agents and facilitating their deployment in practical scenarios.

\section*{Impact Statement}
This work facilitates the real-world deployment of mobile agents, which has been adopted in industry. The trajectory-fusion-based assessment approach is also applicable to the evaluation of other types of agents. This work will contribute to enhancing the practicality of agents in accomplishing various GUI tasks.

\bibliography{main}

@article{rawles2024androidworld,
  title={Androidworld: A dynamic benchmarking environment for autonomous agents},
  author={Rawles, Christopher and Clinckemaillie, Sarah and Chang, Yifan and Waltz, Jonathan and Lau, Gabrielle and Fair, Marybeth and Li, Alice and Bishop, William and Li, Wei and Campbell-Ajala, Folawiyo and others},
  journal={arXiv preprint arXiv:2405.14573},
  year={2024}
}

@misc{ye2025mobileagentv3fundamentalagentsgui,
      title={Mobile-Agent-v3: Fundamental Agents for GUI Automation}, 
      author={Jiabo Ye and Xi Zhang and Haiyang Xu and Haowei Liu and Junyang Wang and Zhaoqing Zhu and Ziwei Zheng and Feiyu Gao and Junjie Cao and Zhengxi Lu and Jitong Liao and Qi Zheng and Fei Huang and Jingren Zhou and Ming Yan},
      year={2025},
      eprint={2508.15144},
      archivePrefix={arXiv},
      primaryClass={cs.AI},
      url={https://arxiv.org/abs/2508.15144}, 
}

@inproceedings{chen2024spa,
  title={Spa-bench: A comprehensive benchmark for smartphone agent evaluation},
  author={Chen, Jingxuan and Yuen, Derek and Xie, Bin and Yang, Yuhao and Chen, Gongwei and Wu, Zhihao and Yixing, Li and Zhou, Xurui and Liu, Weiwen and Wang, Shuai and others},
  booktitle={NeurIPS 2024 Workshop on Open-World Agents},
  year={2024}
}

@article{bu2025limits,
  title={What Limits Virtual Agent Application? OmniBench: A Scalable Multi-Dimensional Benchmark for Essential Virtual Agent Capabilities},
  author={Bu, Wendong and Wu, Yang and Yu, Qifan and Gao, Minghe and Miao, Bingchen and Zhang, Zhenkui and Pan, Kaihang and Li, Yunfei and Li, Mengze and Ji, Wei and others},
  journal={arXiv preprint arXiv:2506.08933},
  year={2025}
}

@article{qin2025ui,
  title={Ui-tars: Pioneering automated gui interaction with native agents},
  author={Qin, Yujia and Ye, Yining and Fang, Junjie and Wang, Haoming and Liang, Shihao and Tian, Shizuo and Zhang, Junda and Li, Jiahao and Li, Yunxin and Huang, Shijue and others},
  journal={arXiv preprint arXiv:2501.12326},
  year={2025}
}

@article{wang2025ui,
  title={Ui-tars-2 technical report: Advancing gui agent with multi-turn reinforcement learning},
  author={Wang, Haoming and Zou, Haoyang and Song, Huatong and Feng, Jiazhan and Fang, Junjie and Lu, Junting and Liu, Longxiang and Luo, Qinyu and Liang, Shihao and Huang, Shijue and others},
  journal={arXiv preprint arXiv:2509.02544},
  year={2025}
}

@article{zhang2025mobiagent,
  title={MobiAgent: A Systematic Framework for Customizable Mobile Agents},
  author={Zhang, Cheng and Feng, Erhu and Zhao, Xi and Zhao, Yisheng and Gong, Wangbo and Sun, Jiahui and Du, Dong and Hua, Zhichao and Xia, Yubin and Chen, Haibo},
  journal={arXiv preprint arXiv:2509.00531},
  year={2025}
}

@article{fabbri2025multinrc,
  title={MultiNRC: A Challenging and Native Multilingual Reasoning Evaluation Benchmark for LLMs},
  author={Fabbri, Alexander R and Mares, Diego and Flores, Jorge and Mankikar, Meher and Hernandez, Ernesto and Lee, Dean and Liu, Bing and Xing, Chen},
  journal={arXiv preprint arXiv:2507.17476},
  year={2025}
}

@article{zheng2023judging,
  title={Judging llm-as-a-judge with mt-bench and chatbot arena},
  author={Zheng, Lianmin and Chiang, Wei-Lin and Sheng, Ying and Zhuang, Siyuan and Wu, Zhanghao and Zhuang, Yonghao and Lin, Zi and Li, Zhuohan and Li, Dacheng and Xing, Eric and others},
  journal={Advances in neural information processing systems},
  volume={36},
  pages={46595--46623},
  year={2023}
}

@article{he2024multi,
  title={Multi-if: Benchmarking llms on multi-turn and multilingual instructions following},
  author={He, Yun and Jin, Di and Wang, Chaoqi and Bi, Chloe and Mandyam, Karishma and Zhang, Hejia and Zhu, Chen and Li, Ning and Xu, Tengyu and Lv, Hongjiang and others},
  journal={arXiv preprint arXiv:2410.15553},
  year={2024}
}

@article{chai2025a3,
  title={A3: Android agent arena for mobile gui agents},
  author={Chai, Yuxiang and Li, Hanhao and Zhang, Jiayu and Liu, Liang and Liu, Guangyi and Wang, Guozhi and Ren, Shuai and Huang, Siyuan and Li, Hongsheng},
  journal={arXiv preprint arXiv:2501.01149},
  year={2025}
}

@article{rawles2023androidinthewild,
  title={Androidinthewild: A large-scale dataset for android device control},
  author={Rawles, Christopher and Li, Alice and Rodriguez, Daniel and Riva, Oriana and Lillicrap, Timothy},
  journal={Advances in Neural Information Processing Systems},
  volume={36},
  pages={59708--59728},
  year={2023}
}

@inproceedings{zhang-etal-2024-android,
    title = "Android in the Zoo: Chain-of-Action-Thought for {GUI} Agents",
    author = "Zhang, Jiwen  and
      Wu, Jihao  and
      Yihua, Teng  and
      Liao, Minghui  and
      Xu, Nuo  and
      Xiao, Xiao  and
      Wei, Zhongyu  and
      Tang, Duyu",
    editor = "Al-Onaizan, Yaser  and
      Bansal, Mohit  and
      Chen, Yun-Nung",
    booktitle = "Findings of the Association for Computational Linguistics: EMNLP 2024",
    month = nov,
    year = "2024",
    address = "Miami, Florida, USA",
    publisher = "Association for Computational Linguistics",
    url = "https://aclanthology.org/2024.findings-emnlp.702/",
    doi = "10.18653/v1/2024.findings-emnlp.702",
    pages = "12016--12031"
}

@article{leung2025androidcontrol,
  title={AndroidControl-Curated: Revealing the True Potential of GUI Agents through Benchmark Purification},
  author={Leung, Ho Fai and Xi, Xiaoyan and Zuo, Fei},
  journal={arXiv preprint arXiv:2510.18488},
  year={2025}
}

@article{xu2025mobile,
  title={Mobile-Bench-v2: A More Realistic and Comprehensive Benchmark for VLM-based Mobile Agents},
  author={Xu, Weikai and Jiang, Zhizheng and Liu, Yuxuan and Gao, Pengzhi and Liu, Wei and Luan, Jian and Li, Yuanchun and Liu, Yunxin and Wang, Bin and An, Bo},
  journal={arXiv preprint arXiv:2505.11891},
  year={2025}
}

@article{zhang2024mobile,
  title={Mobile-env: an evaluation platform and benchmark for LLM-GUI interaction},
  author={Zhang, Danyang and Xu, Hongshen and Zhao, Zihan and Chen, Lu and Cao, Ruisheng and Yu, Kai},
  journal={arXiv preprint arXiv:2305.08144},
  year={2024}
}

@inproceedings{xing2024understanding,
  title={Understanding the weakness of large language model agents within a complex android environment},
  author={Xing, Mingzhe and Zhang, Rongkai and Xue, Hui and Chen, Qi and Yang, Fan and Xiao, Zhen},
  booktitle={Proceedings of the 30th ACM SIGKDD Conference on Knowledge Discovery and Data Mining},
  pages={6061--6072},
  year={2024}
}

@inproceedings{xu2025androidlab,
  title={Androidlab: Training and systematic benchmarking of android autonomous agents},
  author={Xu, Yifan and Liu, Xiao and Sun, Xueqiao and Cheng, Siyi and Yu, Hao and Lai, Hanyu and Zhang, Shudan and Zhang, Dan and Tang, Jie and Dong, Yuxiao},
  booktitle={Proceedings of the 63rd Annual Meeting of the Association for Computational Linguistics (Volume 1: Long Papers)},
  pages={2144--2166},
  year={2025}
}

@misc{lee2025benchmarkingmobiledevicecontrol,
      title={Benchmarking Mobile Device Control Agents across Diverse Configurations}, 
      author={Juyong Lee and Taywon Min and Minyong An and Dongyoon Hahm and Haeone Lee and Changyeon Kim and Kimin Lee},
      year={2025},
      eprint={2404.16660},
      archivePrefix={arXiv},
      primaryClass={cs.HC},
      url={https://arxiv.org/abs/2404.16660}, 
}

@article{song2025colorbench,
  title={ColorBench: Benchmarking Mobile Agents with Graph-Structured Framework for Complex Long-Horizon Tasks},
  author={Song, Yuanyi and Huang, Heyuan and Lin, Qiqiang and Zhao, Yin and Qu, Xiangmou and Wang, Jun and Lou, Xingyu and Liu, Weiwen and Zhang, Zhuosheng and Yu, Yong and others},
  journal={arXiv preprint arXiv:2510.14621},
  year={2025}
}

@article{wang2024mobileagentbench,
  title={Mobileagentbench: An efficient and user-friendly benchmark for mobile llm agents},
  author={Wang, Luyuan and Deng, Yongyu and Zha, Yiwei and Mao, Guodong and Wang, Qinmin and Min, Tianchen and Chen, Wei and Chen, Shoufa},
  journal={arXiv preprint arXiv:2406.08184},
  year={2024}
}

@misc{openai2025conversation,
  title = {Conversation with an AI model [Generative AI chat]},
  author = {{OpenAI}},
  year = {2025},
  howpublished = {Available at: \url{chat.openai.com}},
  note = {Accessed: [Date you accessed the tool]}
}

@misc{Anthropic2025Claude4.1,
  author = {{Anthropic}},
  title = {Claude Opus 4.1},
  howpublished = {Large language model, \url{https://www.anthropic.com}},
  year = {2025},
  note = {Accessed: [Insert Full Date You Accessed the Model, e.g., 6 Jan. 2026]},
}

@article{comanici2025gemini,
  title={Gemini 2.5: Pushing the frontier with advanced reasoning, multimodality, long context, and next generation agentic capabilities},
  author={Comanici, Gheorghe and Bieber, Eric and Schaekermann, Mike and Pasupat, Ice and Sachdeva, Noveen and Dhillon, Inderjit and Blistein, Marcel and Ram, Ori and Zhang, Dan and Rosen, Evan and others},
  journal={arXiv preprint arXiv:2507.06261},
  year={2025}
}

@article{liu2024autoglm,
  title={Autoglm: Autonomous foundation agents for guis},
  author={Liu, Xiao and Qin, Bo and Liang, Dongzhu and Dong, Guang and Lai, Hanyu and Zhang, Hanchen and Zhao, Hanlin and Iong, Iat Long and Sun, Jiadai and Wang, Jiaqi and others},
  journal={arXiv preprint arXiv:2411.00820},
  year={2024}
}

@article{ye2025mobile,
  title={Mobile-agent-v3: Fundamental agents for gui automation},
  author={Ye, Jiabo and Zhang, Xi and Xu, Haiyang and Liu, Haowei and Wang, Junyang and Zhu, Zhaoqing and Zheng, Ziwei and Gao, Feiyu and Cao, Junjie and Lu, Zhengxi and others},
  journal={arXiv preprint arXiv:2508.15144},
  year={2025}
}

@misc{lu2024omniparserpurevisionbased,
      title={OmniParser for Pure Vision Based GUI Agent}, 
      author={Yadong Lu and Jianwei Yang and Yelong Shen and Ahmed Awadallah},
      year={2024},
      eprint={2408.00203},
      archivePrefix={arXiv},
      primaryClass={cs.CV},
      url={https://arxiv.org/abs/2408.00203}, 
}

@inproceedings{kwon2023efficient,
  title={Efficient Memory Management for Large Language Model Serving with PagedAttention},
  author={Woosuk Kwon and Zhuohan Li and Siyuan Zhuang and Ying Sheng and Lianmin Zheng and Cody Hao Yu and Joseph E. Gonzalez and Hao Zhang and Ion Stoica},
  booktitle={Proceedings of the ACM SIGOPS 29th Symposium on Operating Systems Principles},
  year={2023}
}

@article{xie2024osworld,
  title={Osworld: Benchmarking multimodal agents for open-ended tasks in real computer environments},
  author={Xie, Tianbao and Zhang, Danyang and Chen, Jixuan and Li, Xiaochuan and Zhao, Siheng and Cao, Ruisheng and Hua, Toh J and Cheng, Zhoujun and Shin, Dongchan and Lei, Fangyu and others},
  journal={Advances in Neural Information Processing Systems},
  volume={37},
  pages={52040--52094},
  year={2024}
}

@article{ma2024coco,
  title={Coco-agent: A comprehensive cognitive mllm agent for smartphone gui automation},
  author={Ma, Xinbei and Zhang, Zhuosheng and Zhao, Hai},
  journal={arXiv preprint arXiv:2402.11941},
  year={2024}
}

@article{zhang2024large,
  title={Large language model-brained gui agents: A survey},
  author={Zhang, Chaoyun and He, Shilin and Qian, Jiaxu and Li, Bowen and Li, Liqun and Qin, Si and Kang, Yu and Ma, Minghua and Liu, Guyue and Lin, Qingwei and others},
  journal={arXiv preprint arXiv:2411.18279},
  year={2024}
}

@article{toyama2021androidenv,
  title={Androidenv: A reinforcement learning platform for android},
  author={Toyama, Daniel and Hamel, Philippe and Gergely, Anita and Comanici, Gheorghe and Glaese, Amelia and Ahmed, Zafarali and Jackson, Tyler and Mourad, Shibl and Precup, Doina},
  journal={arXiv preprint arXiv:2105.13231},
  year={2021}
}

@inproceedings{chai-etal-2025-amex,
    title = "{AMEX}: Android Multi-annotation Expo Dataset for Mobile {GUI} Agents",
    author = "Chai, Yuxiang  and
      Huang, Siyuan  and
      Niu, Yazhe  and
      Xiao, Han  and
      Liu, Liang  and
      Wang, Guozhi  and
      Zhang, Dingyu  and
      Ren, Shuai  and
      Li, Hongsheng",
    editor = "Che, Wanxiang  and
      Nabende, Joyce  and
      Shutova, Ekaterina  and
      Pilehvar, Mohammad Taher",
    booktitle = "Findings of the Association for Computational Linguistics: ACL 2025",
    month = jul,
    year = "2025",
    address = "Vienna, Austria",
    publisher = "Association for Computational Linguistics",
    url = "https://aclanthology.org/2025.findings-acl.110/",
    doi = "10.18653/v1/2025.findings-acl.110",
    pages = "2138--2156",
    ISBN = "979-8-89176-256-5"
}

@article{lee2024benchmarking,
  title={Benchmarking mobile device control agents across diverse configurations},
  author={Lee, Juyong and Min, Taywon and An, Minyong and Hahm, Dongyoon and Lee, Haeone and Kim, Changyeon and Lee, Kimin},
  journal={arXiv preprint arXiv:2404.16660},
  year={2024}
}

@article{kong2025mobileworld,
  title={MobileWorld: Benchmarking Autonomous Mobile Agents in Agent-User Interactive, and MCP-Augmented Environments},
  author={Kong, Quyu and Zhang, Xu and Yang, Zhenyu and Gao, Nolan and Liu, Chen and Tong, Panrong and Cai, Chenglin and Zhou, Hanzhang and Zhang, Jianan and Chen, Liangyu and others},
  journal={arXiv preprint arXiv:2512.19432},
  year={2025}
}

@article{yan2025mcpworld,
  title={MCPWorld: A Unified Benchmarking Testbed for API, GUI, and Hybrid Computer Use Agents},
  author={Yan, Yunhe and Wang, Shihe and Du, Jiajun and Yang, Yexuan and Shan, Yuxuan and Qiu, Qichen and Jia, Xianqing and Wang, Xinge and Yuan, Xin and Han, Xu and others},
  journal={arXiv preprint arXiv:2506.07672},
  year={2025}
}
\bibliographystyle{icml2026}

\newpage
\appendix

\onecolumn %
\section{Real-world Trajectory Collection}
\label{app:collect}
\begin{figure}[h]
  \centering 
  \includegraphics[width=0.9\columnwidth]{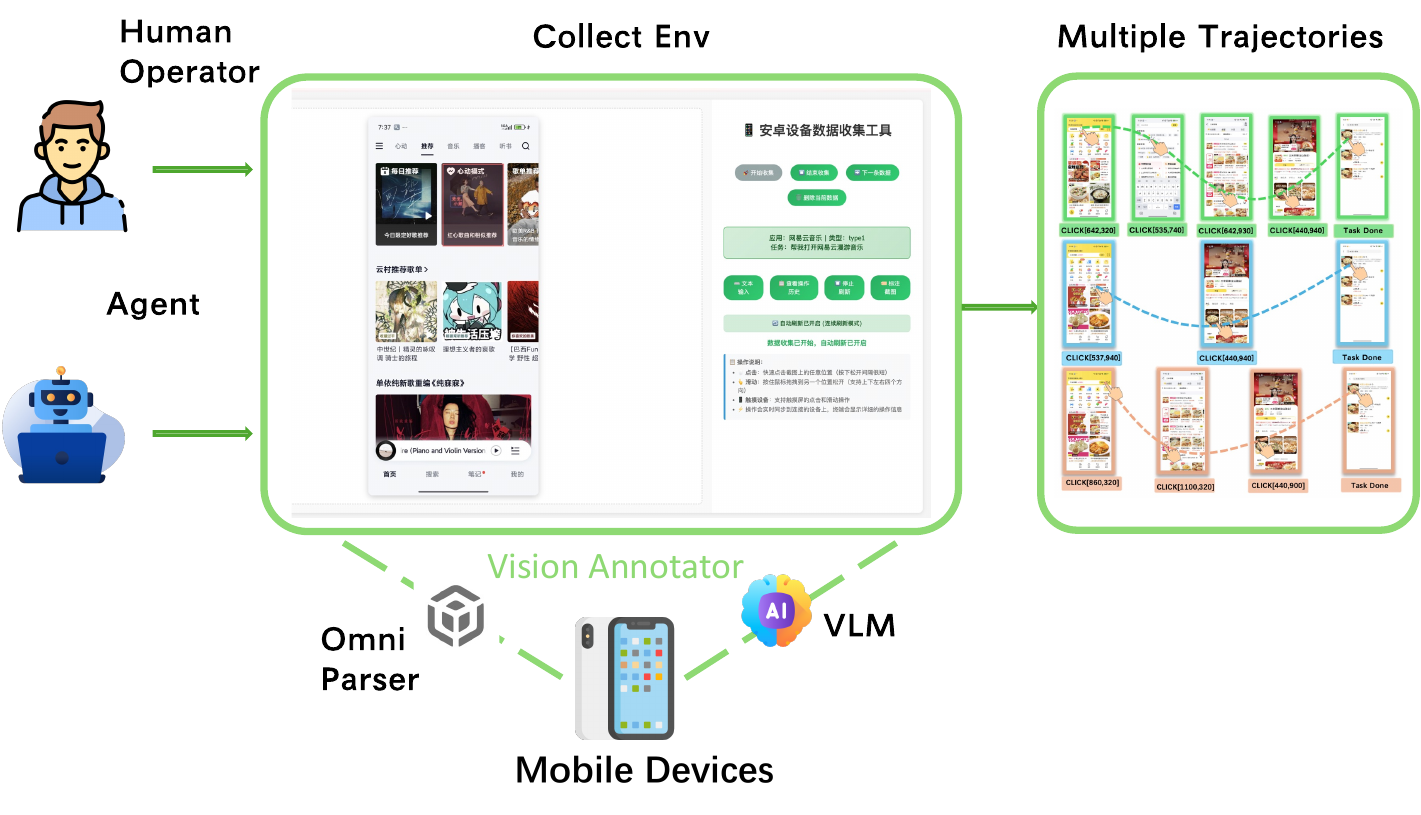}
  \caption{
    \textbf{Efficient trajectory collection tools}. A trajectory collection framework based on real devices and actual applications.
  }
  \label{fig:collect}
\end{figure}

To ensure the collected trajectories accurately reflect real-world user behavior, we have developed a lightweight action recording tool for smartphones(see fig\ref{fig:collect}). During data collection, annotators or agents interact with this tool, which captures and logs every user operation before forwarding it to the device. The tool renders bounding boxes for all interactive on-screen elements based on XML files that describe UI hierarchies. If a bounding box is missing due to incomplete XML data, OmniParser is employed to regenerate the corresponding bounding box. The resulting trajectories can subsequently be used for graph construction, evaluation, and other follow-up steps.

Within the graph construction algorithm, the labeling of observed pages can be performed concurrently. Following predefined labeling rules (which may be specified based on the transition structure of observed interfaces), multiple trajectories are merged into a unified state transition graph. Furthermore, to extend the state transition graph, it suffices to append new trajectories according to the same labeling principles, thereby ensuring strong extensibility.

\section{Action Space}
\label{app:act_space}
This section presents the action space supported by our environment, which includes operations such as click, swipe, wait, text input, back, and more.
\begin{table}[H]
\centering
\small
\begin{tabular}{l | c | c}

Action & Parameter & Description \\
\midrule
\multicolumn{3}{c}{\textbf{GUI Operation}} \\
\midrule
click & x,y & Click at the specified coordinates \\
input & string & Type text into the focused field \\
swipe & start\_x,start\_y,end\_x,end\_y & Swipe from start to end coordinates \\
wait & -- & Wait for update \\
double\_click & x,y &Double Click at the specified coordinates \\
long\_press & x,y & Long press at the specified coordinates \\
\midrule
\multicolumn{3}{c}{\textbf{System Operation}} \\
\midrule
back & -- & Navigate to the previous screen \\
home & -- & Navigate to the Home screen \\
\midrule
\multicolumn{3}{c}{\textbf{Task Control Operation}} \\  
\midrule
done & -- & Task done \\
failed & -- & Task failed \\
\bottomrule
\end{tabular}
\caption{Action types and their parameters in the task execution framework.}  
\label{tab:action_types}
\vspace{-0.1in}
\end{table}

\section{Task Graph Construction Details}
\label{app:graph_build}

This section introduces the details of the construction. The task graph primarily consists of two elements: node construction and edge construction. For node construction, we provide the following data structure, which describes the attributes of a state, including observation information (State\_Info), labeling properties (Tag), and state transition mappings(Transition\_Space).

\begin{lstlisting}[
    frame=lines,
    framesep=2mm,
    numbersep=5pt,
    language=,
    basicstyle=\ttfamily\small
]
@dataclass
class State:
    '''
    Task Transition Graph state 
    '''
    State_Info      # screenshots or other (e.g., text or XML)
    Tag             # State tagging
    
    # Action-space-based interface transition hashing
    Transition_Space = {
        "click": {      # Box[x1,y1,x2,y2] --> next state
        },
        "swipe": {       # (Direction,Distance) --> next state
        },
        "input": {       # Text --> next state
        },
        "wait": {        # Duration --> next state
        },
        "back": {        # --> prev state
        },
        "home": {        # --> home state
        },
        "long Press": {  # Box[x1,y1,x2,y2] --> next state
        } 
        # extend...
    }
\end{lstlisting}

For edge construction, we first traverse the raw trajectories to assign state-specific labels and record all observed action transitions, thereby populating the transition space of each abstract state. The labels can be manually defined based on the structural distribution of UI elements on the interface or inferred by a model; the key requirement is that states sharing the same label can reuse a common transition space. Specifically, each raw observation is mapped to an abstract state via semantic labeling or hashing, and directed edges are added according to the corresponding $(u, a, v)$ transitions. Subsequently, nodes with identical semantic labels are merged by taking the union of their transition spaces, aggregating historical actions across equivalent states to maximize action coverage while reducing state fragmentation.

\begin{algorithm}[H]
   \caption{Offline Graph Construction via Trajectory Merging}
   \label{alg:construction}
\begin{algorithmic}[1]
   \STATE {\bfseries Input:} Set of raw trajectories $\mathcal{D} = \{ \tau_1, \dots, \tau_N \}$, where $\tau_i = \{(o_1, a_1), \dots, (o_T, \cdot)\}$
   \STATE {\bfseries Output:} State Transition Graph $\mathcal{G} = (\mathcal{V}, \mathcal{E})$
   
   \vspace{0.1cm}
   \STATE \textit{// Core Logic Overview:}
   \STATE \textit{// 1. Abstraction: Map raw UI observations to nodes via Semantic Labels or Hashing.}
   \STATE \textit{// 2. Construction: Build directed edges based on recorded action transitions.}
   \STATE \textit{// 3. Unification: Merge action spaces of identical states to maximize coverage.}
   \vspace{0.1cm}

   \STATE Initialize node set $\mathcal{V} \leftarrow \emptyset$, edge map $\mathcal{E} \leftarrow \emptyset$
   \STATE Define abstraction function $\Phi(o) \to s_{id}$
   
   \STATE \COMMENT{\textbf{Phase 1: Trajectory Traversal \& Topology Building}}
   \FOR{each trajectory $\tau \in \mathcal{D}$}
      \FOR{each transition step $(o_t, a_t, o_{t+1})$ in $\tau$}
         \STATE \COMMENT{Step 1: State Abstraction}
         \IF{$o_t$ has semantic label $L$}
            \STATE $u \leftarrow L$ 
         \ELSE
            \STATE $u \leftarrow \textsc{Hash}(o_t)$ 
         \ENDIF
         \STATE $v \leftarrow \Phi(o_{t+1})$ 
         
         \STATE \COMMENT{Step 2: Graph Update}
         \STATE $\mathcal{V} \leftarrow \mathcal{V} \cup \{u, v\}$
         \IF{$u \notin \mathcal{E}$ or $a_t \notin \mathcal{E}[u]$}
            \STATE $\mathcal{E}[u][a_t] \leftarrow v$ 
         \ELSE
            \STATE $\mathcal{E}[u][a_t] \leftarrow \mathcal{E}[u][a_t] \cup \{v\}$ 
         \ENDIF
      \ENDFOR
   \ENDFOR
   
   \STATE \COMMENT{\textbf{Phase 2: Transition Space Unification}}
   \FOR{each abstract state $u \in \mathcal{V}$}
      \IF{$u$ is a labeled node}
         \STATE $\mathcal{T}_{all} \leftarrow \bigcup_{s \in \Phi^{-1}(u)} \text{Transitions}(s)$
         \STATE $\mathcal{E}[u] \leftarrow \mathcal{E}[u] \cup \mathcal{T}_{all}$ \COMMENT{Aggregate all historical actions}
      \ENDIF
   \ENDFOR

   \STATE \textbf{return} $\mathcal{G} = (\mathcal{V}, \mathcal{E})$
\end{algorithmic}
\end{algorithm}

\section{Statistics on the Number of Interface Actions}
\label{app:statistic}
\begin{figure}[H]
  \begin{center}
    \centerline{\includegraphics[width=\columnwidth]{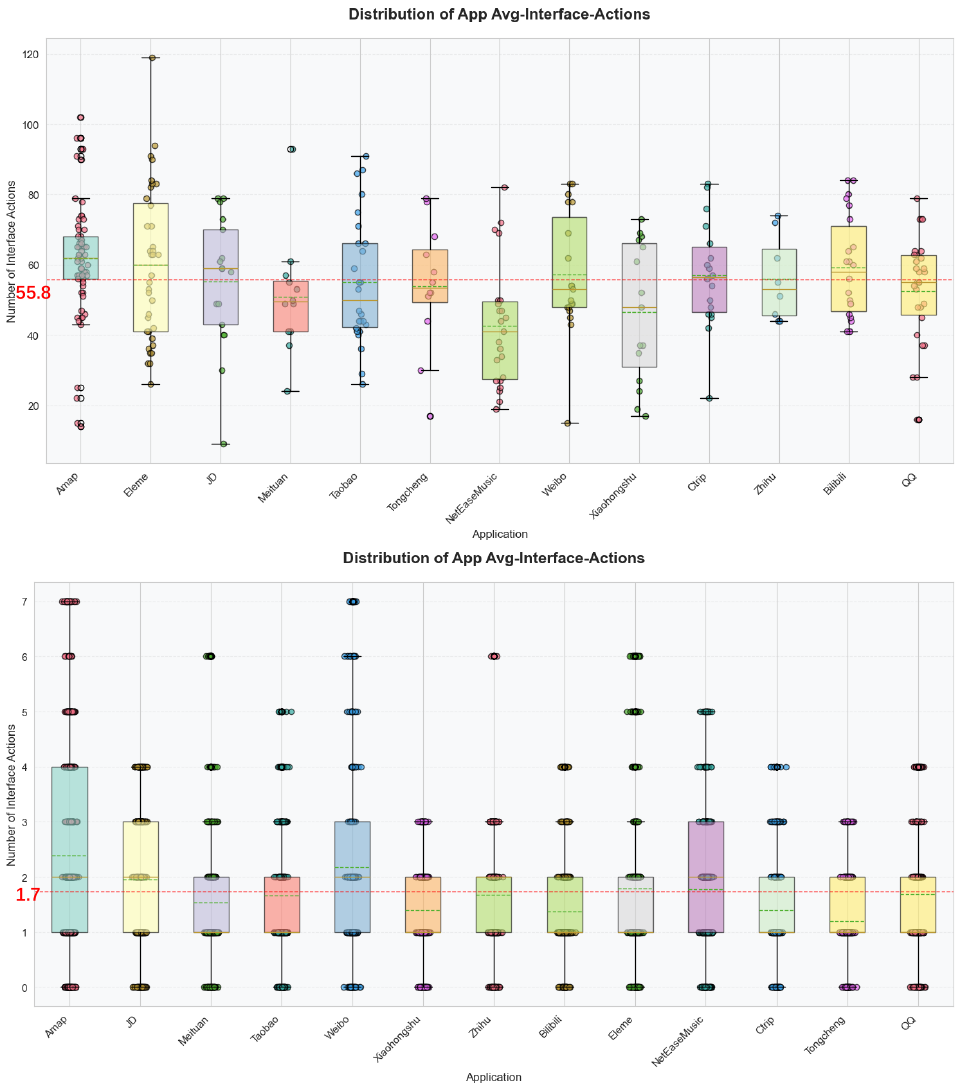}}
    \caption{
      This figure illustrates the statistical distribution of transition action counts on individual interfaces across different mobile applications(Due to space limitations, we present only a subset of the statistical results).
    }
    \label{fig:actnums}
  \end{center}
\end{figure}

We employ OminiParserV2 to conduct a statistical analysis of icon elements corresponding to executable actions across various interface types in multiple mainstream applications. Duplicate elements are removed using non-maximum suppression, and element independence is ensured through intersection detection. The resulting statistics are presented in the upper part of Figure\ref{fig:actnums}. For multiple feasible trajectories that accomplish the same task, we further compute the average number of navigation actions per interface, with the results shown in the lower part of Figure\ref{fig:actnums}.

\section{State transition graph visualization}
\label{app:graph_vis}
Based on our graph construction algorithm, which merges multiple trajectories into a task state transition graph, we randomly selected four tasks for visualization. The visualization results are presented below.
\begin{figure}[H]
  \vskip 0.2in
  \begin{center}
    \centerline{\includegraphics[width=\columnwidth]{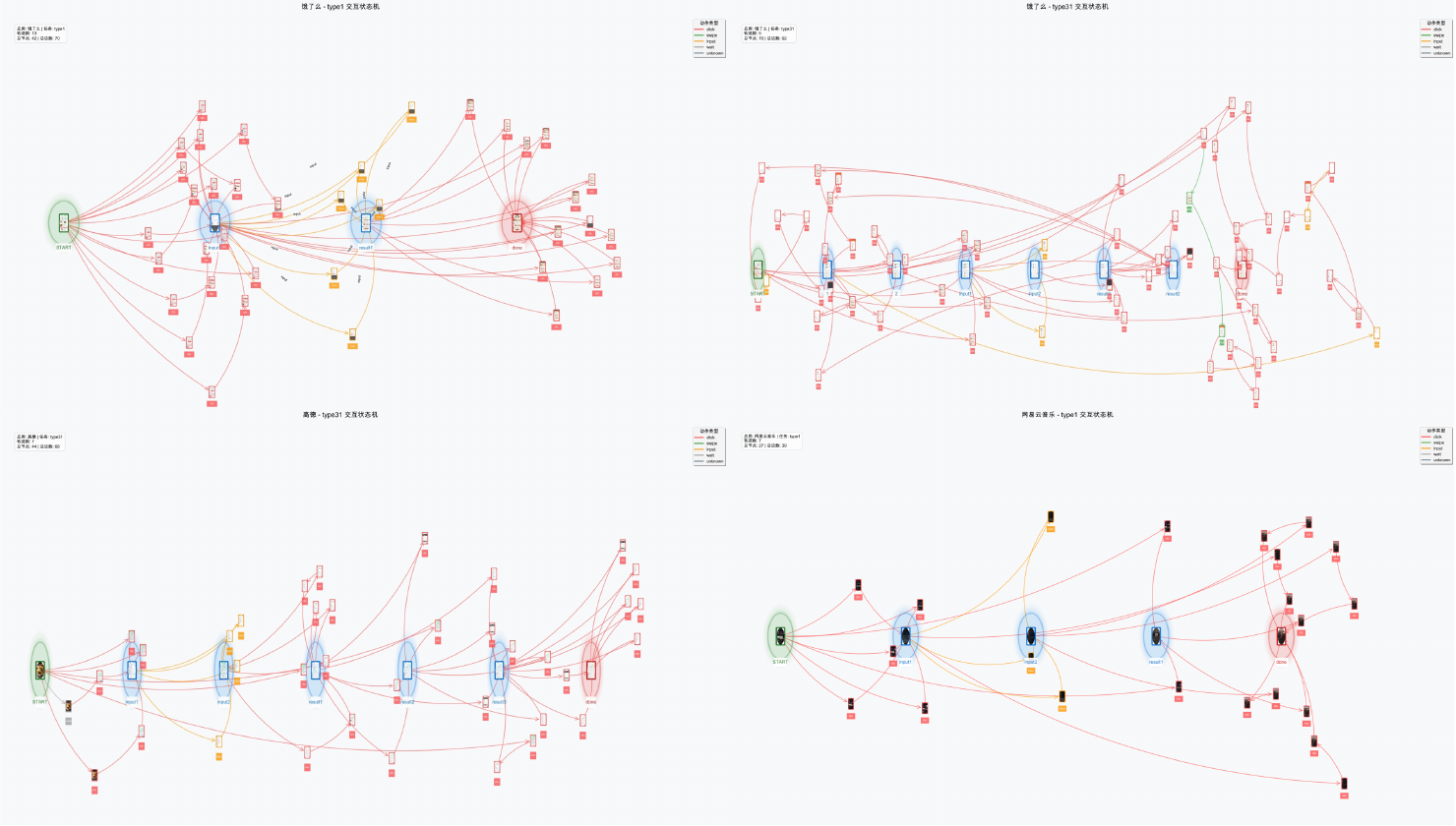}}
    \caption{
      \textbf{Visualization Examples of Task State Transition Graphs}, including Different Tasks from Ele.me, AutoNavi, and NetEase Cloud Music Apps.
    }
    \label{fig:graph_vis}
  \end{center}
\end{figure}

\section{Execution Details of General Models}
\label{app:model_exec}

\begin{tcolorbox}[title=System Prompt,breakable]
\textbf{Role Definition}
You are a mobile phone operation AI assistant, tasked with helping the user complete the following task: \texttt{"\{task\_description\}"}.

\textbf{Input Description}
I will provide you with:
\begin{enumerate}
    \item \textbf{Action History}: A record of all previous operations.
    \item \textbf{Screen Screenshot}: A complete screenshot of the current phone screen.
    \item \textbf{\{layer\_count\} Annotated Screenshots}: Annotation layers of clickable elements generated based on the screen screenshot. To avoid element overlap, all clickable elements are distributed across different layers for display. The union of elements across all layers represents the complete set of clickable elements.
    \begin{itemize}
        \item Clickable elements are marked with red bounding boxes.
        \item Each element's index is displayed inside the top-left corner of its red bounding box as a red-background white-digit number.
    \end{itemize}
\end{enumerate}

\textbf{Action History}
\texttt{\{history\}}

\textbf{Task Requirements}
Please carefully analyze the current screen state and action history, then determine the most appropriate next action.

\textbf{Actions Available}
\begin{enumerate}
    \item \textbf{Click (click)}
    \begin{itemize}
        \item Parameters: \texttt{index} (integer, corresponding to the UI element index in the annotated screenshots)
        \item Parameters: \texttt{target\_element} (string, describing the UI element to be clicked)
        \item \textbf{Critical}: You must carefully observe the annotated screenshots, find the red bounding box that best matches the description of the \texttt{target\_element}, and use the number displayed inside the top-left corner of that box (red background, white digits) as the \texttt{index}.
        \item \textbf{Prevent Mis-selection}: The reasoning must explicitly explain why this specific red bounding box was chosen and not an adjacent one.
    \end{itemize}

    \item \textbf{Swipe (swipe)}
    \begin{itemize}
        \item Parameters: \texttt{direction} (string, must be one of: UP, DOWN, LEFT, RIGHT)
        \item \textbf{Critical}: Direction Clarification: UP means swiping the finger upward to scroll content upward, revealing content below; DOWN means swiping downward to scroll content downward, revealing content above; LEFT means swiping left to scroll content left; RIGHT means swiping right to scroll content right.
    \end{itemize}

    \item \textbf{double click}
    \begin{itemize}
        \item Parameters: \texttt{index} (integer, corresponding to the UI element index in the annotated screenshots) 
    \end{itemize}

    \item \textbf{long press}
    \begin{itemize}
        \item Parameters: \texttt{index} (integer, corresponding to the UI element index in the annotated screenshots) 
    \end{itemize}

    \item \textbf{Home}
    \begin{itemize}
        \item No parameters, navigate to the Home screen.    
    \end{itemize}

    \item \textbf{Home}
    \begin{itemize}
        \item No parameters, navigate to the Home screen.    
    \end{itemize}

    \item \textbf{Home}
    \begin{itemize}
        \item No parameters, navigate to the Home screen.    
    \end{itemize}

    \item \textbf{Back (back)}
    \begin{itemize}
        \item No parameters, indicates returning to the previous state.
    \end{itemize}

    \item \textbf{Task Complete (done)}
    \begin{itemize}
        \item No parameters, indicates the task is completed.
    \end{itemize}
\end{enumerate}

\textbf{Output Format}
Please output strictly in the following JSON format:
\begin{verbatim}
{
    "reasoning": 
    "Provide a detailed explanation of your analysis and the reason for 
    choosing this action. For click actions, it must include 
    the following complete process:
    1) Detailed description of the target element: including element content, 
    color, shape, size, and other visual features.
    2) Precise location of the target element: its specific position on the 
    screen, using surrounding elements as reference.
    3) Annotation map search process: state in which annotation map (layer 
    number) the matching red bounding box was found.
    4) Red bounding box verification: confirm that the position, contained 
    content, and boundaries of this red box perfectly 
    match the target element.
    5) Index reading confirmation: explicitly state the number found inside 
    the top-left corner of the selected red bounding box.
    6) Final confirmation: reiterate that this choice is correct and no 
    adjacent element was mistakenly selected.
    For input actions, you must first explicitly state:
    1) Whether a soft keyboard is currently displayed on the screen.
    2) Whether the target input field is already activated (has a cursor 
    or is highlighted).
    3) If either of the above checks is negative, you MUST choose a click 
    action first to activate the input field, NOT an input action.",
    "action": "action_name(click/swipe/input/back/done)",
    "parameters": {
        "parameter_name": "parameter_value"
    }
}
\end{verbatim}

\vspace{0.8em}
\textbf{Critical Steps for Index Selection (Mandatory Reading for Click Actions)}

\textbf{Step 1: Precisely describe the visual features of the target element.}
\begin{itemize}
    \item Describe in detail the content, color, shape, and other visual features of the target element.
    \item Precisely describe the element's position on the screen (e.g., top third of the screen, left edge, bottom-right corner).
    \item Describe other UI elements surrounding the target element as reference points.
    \item Example: "Need to click the white input box with the text 'Search', located at the very top of the screen, just below the app title."
\end{itemize}

\textbf{Step 2: Systematically search for the red bounding box.}
\begin{itemize}
    \item \textbf{You MUST examine each annotation map sequentially, one by one.} Do not skip any map.
    \item For each map, first observe the overall distribution of all red bounding boxes.
    \item Focus on finding the red bounding box whose position and features perfectly match the description from Step 1.
    \item \textbf{Critical Requirement: The red bounding box must completely enclose the target element, with boundaries snugly fitting.}
\end{itemize}

\textbf{Step 3: Multi-level verification to ensure correct selection (Most Important Step).}
\begin{itemize}
    \item \textbf{Position Verification}: Confirm the red bounding box's location matches the position described in Step 1 exactly.
    \item \textbf{Content Verification}: Carefully observe whether the content inside the red bounding box is indeed the target element.
    \item \textbf{Boundary Verification}: The boundaries of the red bounding box should be snug against the target element, not containing excessive blank space.
    \item \textbf{Exclude Interference}: If there are multiple similar red boxes, you MUST choose the one whose position matches most precisely.
    \item \textbf{Avoid Adjacent Selection}: Absolutely DO NOT select a red box next to or near the target element.
\end{itemize}

\textbf{Step 4: Read the index number (Final confirmation before execution).}
\begin{itemize}
    \item Re-confirm that the selected red bounding box indeed encloses the correct target element.
    \item Look at the number inside the \textbf{top-left corner} of this red bounding box.
    \item \textbf{Strict Requirement: Must be the top-left corner; the number must be clearly visible.}
    \item This number is the \texttt{index} value to use.
\end{itemize}

\textbf{Step 5: Final Verification.}
\begin{itemize}
    \item In the reasoning, explicitly state: "The red bounding box I selected is located at [specific position], contains [specific content], and its top-left number is [X]."
    \item If you have any uncertainty about the selection, you MUST restart from Step 1.
\end{itemize}

\vspace{0.8em}
\textbf{Critical Steps for Text Input (Mandatory Reading for Input Actions)}

\textbf{Important Prerequisite: Absolutely DO NOT use the input action when the input field is not activated!}

\textbf{Step 1: Mandatory Check of Soft Keyboard Status (Must execute, cannot skip).}
\begin{itemize}
    \item \textbf{Must Check}: Carefully observe if a soft keyboard (virtual keyboard interface) is currently displayed at the very bottom of the screen.
    \item \textbf{Judgment Criterion}: If the bottom of the screen does NOT display a soft keyboard interface containing letter and number keys, it means no input field is activated.
    \item \textbf{Key Rule}: \textbf{You are only allowed to perform an input action when the soft keyboard is fully displayed at the bottom of the screen.}
    \item \textbf{Must State in Reasoning}: "Check soft keyboard status: [Displayed/Not Displayed]."
\end{itemize}

\textbf{Step 2: Activating the Input Field (Must execute if Step 1 check fails).}
\begin{itemize}
    \item \textbf{Strictly Forbidden}: If there is no soft keyboard or the input field is not activated, you absolutely CANNOT use the input action.
    \item \textbf{Must Do}: You MUST first use a click action on the target input field to activate it.
    \item \textbf{Must State in Reasoning}: "Soft keyboard not displayed / Input field not activated. Must click to activate the input field first."
\end{itemize}

\textbf{Step 3: Handling Existing Content.}
\begin{itemize}
    \item If the input field has default text, you may try to clear it or overwrite it directly.
    \item Choose the handling method based on the specific situation.
\end{itemize}

\textbf{Step 4: Execute Text Input (Only when preconditions are satisfied).}
\begin{itemize}
    \item After confirming the soft keyboard is displayed AND the input field is activated, you may use the input action.
    \item After inputting, check that the text in the input field is correct, ensuring there are no input errors, omissions, or extra characters.
    \item In reasoning, you MUST explicitly state: "Confirmed soft keyboard is displayed and input field is activated."
\end{itemize}

\textbf{Step 5: Handling Soft Keyboard After Input (Important).}
\begin{itemize}
    \item \textbf{Must Check After Input}: Observe the key types on the soft keyboard.
    \item \textbf{Criteria for Hiding Soft Keyboard}:
    \begin{itemize}
        \item If the soft keyboard has action buttons like "Search", "OK", "Done", "Send", etc., you should click these buttons.
        \item If the soft keyboard only has navigation buttons like "Next" or "Enter", AND the keyboard is blocking important UI elements, you should click the "down arrow" button (usually top-right) on the soft keyboard to hide it.
    \end{itemize}
    \item \textbf{Must State in Reasoning}: "Check soft keyboard key types: [Action type/Navigation type]. Is the soft keyboard blocking important elements: [Yes/No]. Decision: [Click action button / Hide soft keyboard / Keep as is]."
\end{itemize}

\vspace{0.8em}
\textbf{Strictly Forbidden Input Action Patterns}
\begin{enumerate}
    \item Inputting directly without a soft keyboard.
    \item Inputting directly without describing the check process in reasoning.
    \item Inputting directly upon seeing an input field (must check activation status first).
    \item Not considering soft keyboard obstruction after input.
\end{enumerate}

\vspace{0.5em}
\textbf{The Only Correct Input Action Pattern}
\begin{itemize}
    \item Reasoning includes: "Check soft keyboard status: Displayed. Check input field status: Activated. Confirmed text input is permissible."
    \item Only reasoning containing this complete check process allows the use of the input action.
    \item \textbf{Post-Input Handling}: After input, you must check soft keyboard key types and whether it blocks important elements to decide if it needs to be hidden.
\end{itemize}

\vspace{0.8em}
\textbf{Important Rules}
\begin{enumerate}
    \item \textbf{Position Match Priority}: First determine the element's precise location in the original screenshot, then find the corresponding red bounding box in the annotation maps.
    \item \textbf{Accurate Number Reading}: The \texttt{index} must be the actual number displayed inside the top-left corner of the red bounding box (red background, white digits).
    \item \textbf{Avoid Mis-selecting Adjacent Elements}: This is the most common mistake! Ensure the chosen red bounding box fully encloses the target element, not a nearby similar element.
    \item \textbf{Mandatory Adjacent Element Exclusion Check}: Before selecting any index, you must explicitly explain why other red bounding boxes in the vicinity were NOT chosen.
    \item \textbf{Soft Keyboard Obstruction Handling}: After input, if the soft keyboard is blocking important elements and there is no action button, click the top-right down arrow to hide it.
    \item \textbf{Multi-step Operations}: For complex selections (like date ranges, time slots, cascading options), multiple consecutive actions are required.
    \item \textbf{Special Attention for Date Selection}:
    \begin{itemize}
        \item On a date selection interface, you must first confirm if the currently displayed month is correct.
        \item Do not just click an identical date number; you must ensure the month matches the task requirement.
        \item If the month is wrong, you need to switch to the correct month first, then select the date.
    \end{itemize}
    \item \textbf{Task Completion Judgment}: Use the 'done' action only when the specified task is indeed completed.
    \item \textbf{Operation Coherence}: Each action should be a logical choice based on the current screen state and task goal.
    \item \textbf{Page Error Handling}: If you encounter an incorrect page or a loading failure, you can try going back to the previous level (via a swipe gesture from the leftmost screen edge or by clicking a back button).
\end{enumerate}

\vspace{0.8em}
\textbf{Index Selection Examples}

\textbf{Incorrect Example 1:}
\begin{itemize}
    \item reasoning: "Need to click the search button."
    \item Problem: No description of the element's specific location or visual features.
\end{itemize}

\textbf{Incorrect Example 2:}
\begin{itemize}
    \item reasoning: "Need to click the search box, located at the top of the screen. Found the search box in the annotation map, choose number 8."
    \item Problem: Description is too vague, lacks verification process, prone to selecting the wrong adjacent element.
\end{itemize}

\vspace{0.5em}
\textbf{Correct Example:}
\begin{verbatim}
reasoning: "1) Target element detailed description: Need to click the 
white input box with the placeholder text 'Search', rectangular in shape, 
with a light grey border.
2) Precise location description: This search box is located at the very 
top of the screen, approximately 50 pixels below the status bar, occupies 
about 80% of the screen width, centered.
3) Annotation map search: In annotation map #2, I found a red bounding 
box at the central top position of the screen.
4) Red bounding box verification: This red box completely encloses the 
search input box, its boundaries perfectly align with the edges of the 
input box, and it indeed contains the white input box with the text 
'Search'.
5) Index reading: The top-left inner corner of this red box clearly 
shows the number '15'.
6) Final confirmation: Confirmed this box does not contain any irrelevant 
elements, nor is it an adjacent UI element; it is precisely the search 
box I intend to click."
parameters: {"index": 15, "target_element": "Search input box"}
\end{verbatim}

\end{tcolorbox}
The above constitutes the system prompt designed for task completion by general-purpose models. It defines a constrained action space, provides historical context, and includes both positive and negative examples for in-context learning. Additionally, a style constraint is imposed to require the model to output its reasoning process. Since general-purpose models struggle with generating precise coordinates, we have integrated an auxiliary enhancement tool. This tool utilizes an icon recognition model to overlay bounding boxes and numerical labels directly on the interface image, enabling the model to select from these annotated elements, thereby enhancing its task completion capability.
\begin{figure}[H]
  \begin{center}
    \centerline{\includegraphics[width=\columnwidth]{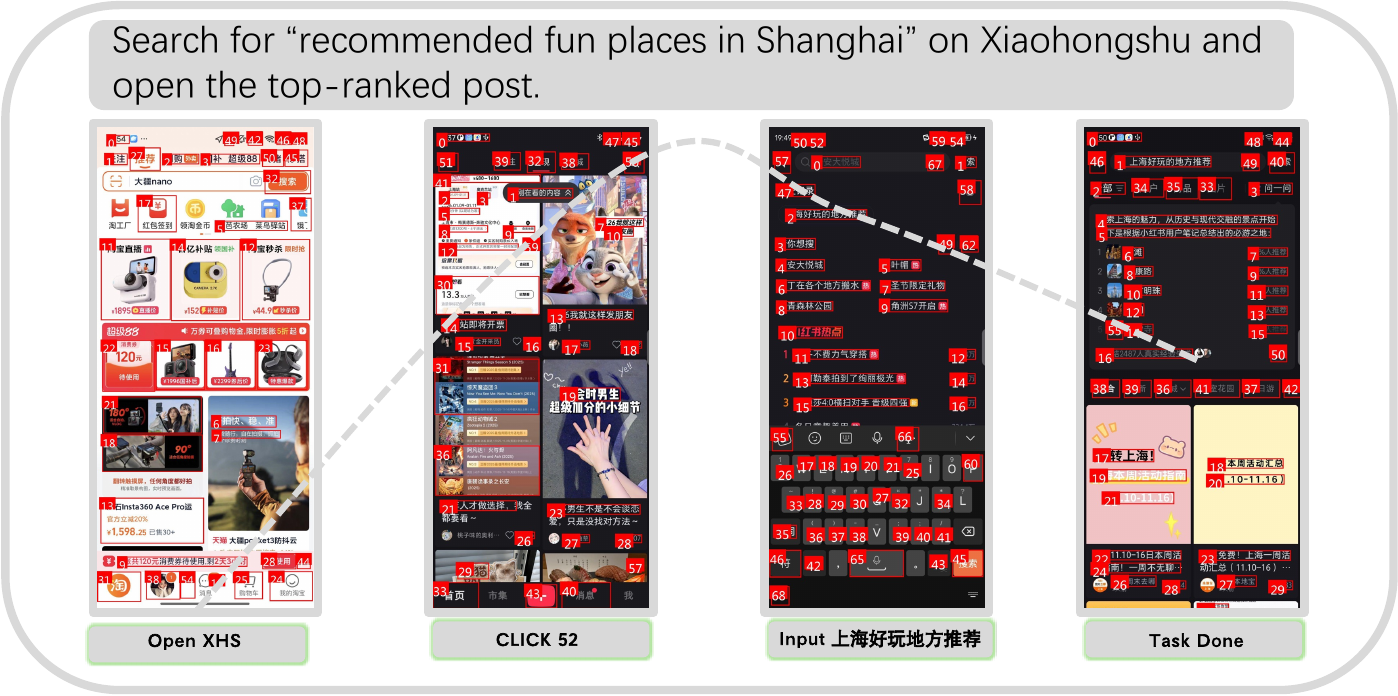}}
    \caption{
     \textbf{Execution Example of General Models}: General models leverage icon recognition models for enhancement.
    }
    \label{fig:gmd}
  \end{center}
\end{figure}

\section{Generalization vs. Accuracy}
\label{app:tree}
\begin{figure}[H]
    \centerline{\includegraphics[width=\columnwidth]{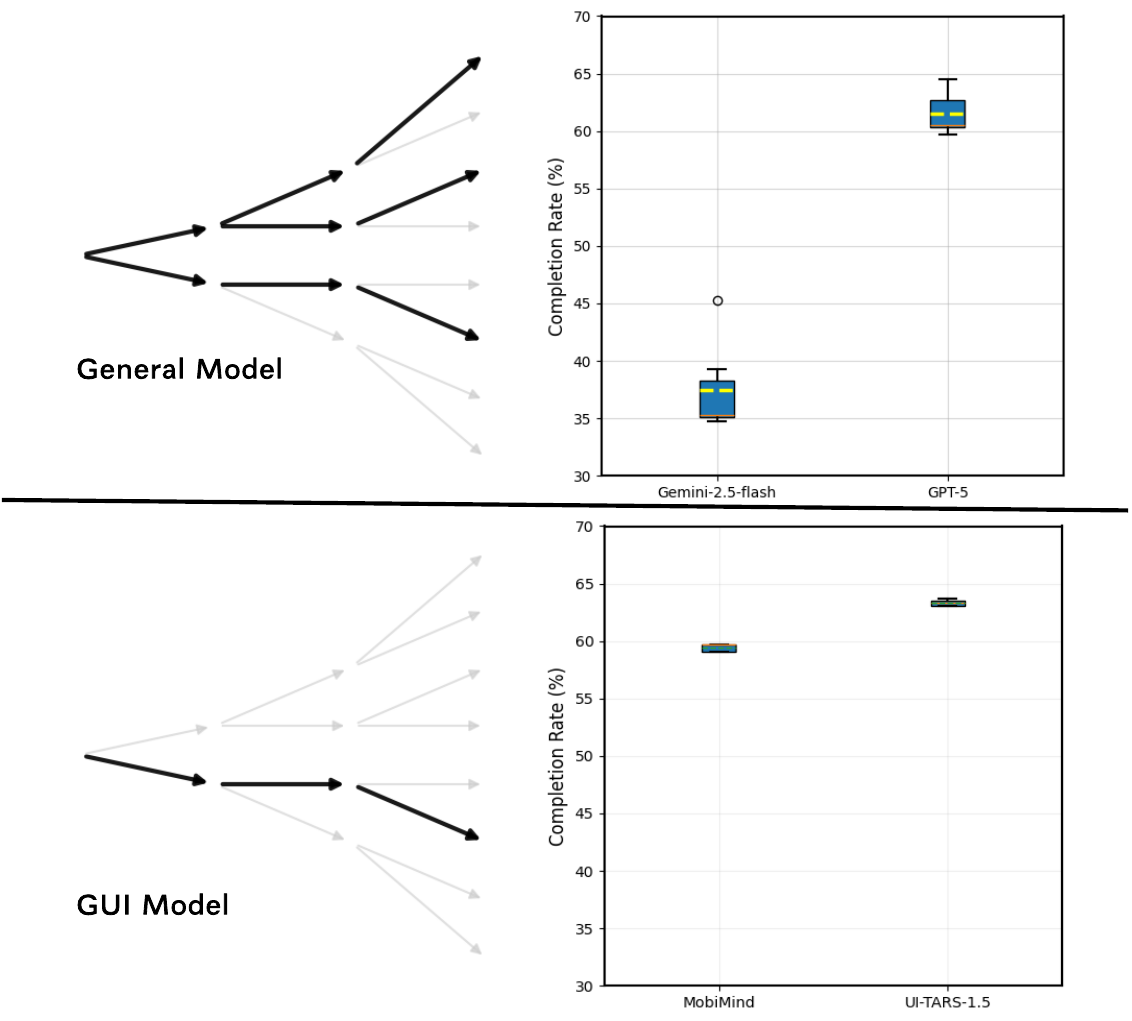}}
    \caption{
      \textbf{The tension between generalization capability and determinism}. The upper figure presents the performance scores of a general-purpose model across multiple sampling trials on randomly selected tasks. The lower figure shows the sampling results of a GUI-specialized model on the same set of tasks.
    }
    \label{fig:tree}
\end{figure}
We selected different general-purpose models (GPT-5, Gemini-2.5-flash) and specialized models (UI-TARS-1.5, MobiMInd), randomly chose multiple tasks, and conducted repeated evaluations under identical configurations. The results revealed that the Completion rate (CR) of general-purpose models exhibited significantly greater fluctuations than that of specialized models. The specialized models produced stable outcomes with consistent execution paths, whereas general-purpose models demonstrated more diverse approaches to task completion.

\section{Resolution Scaling}
\label{app:rf}
\begin{figure}[H]
    \centerline{\includegraphics[width=\columnwidth]{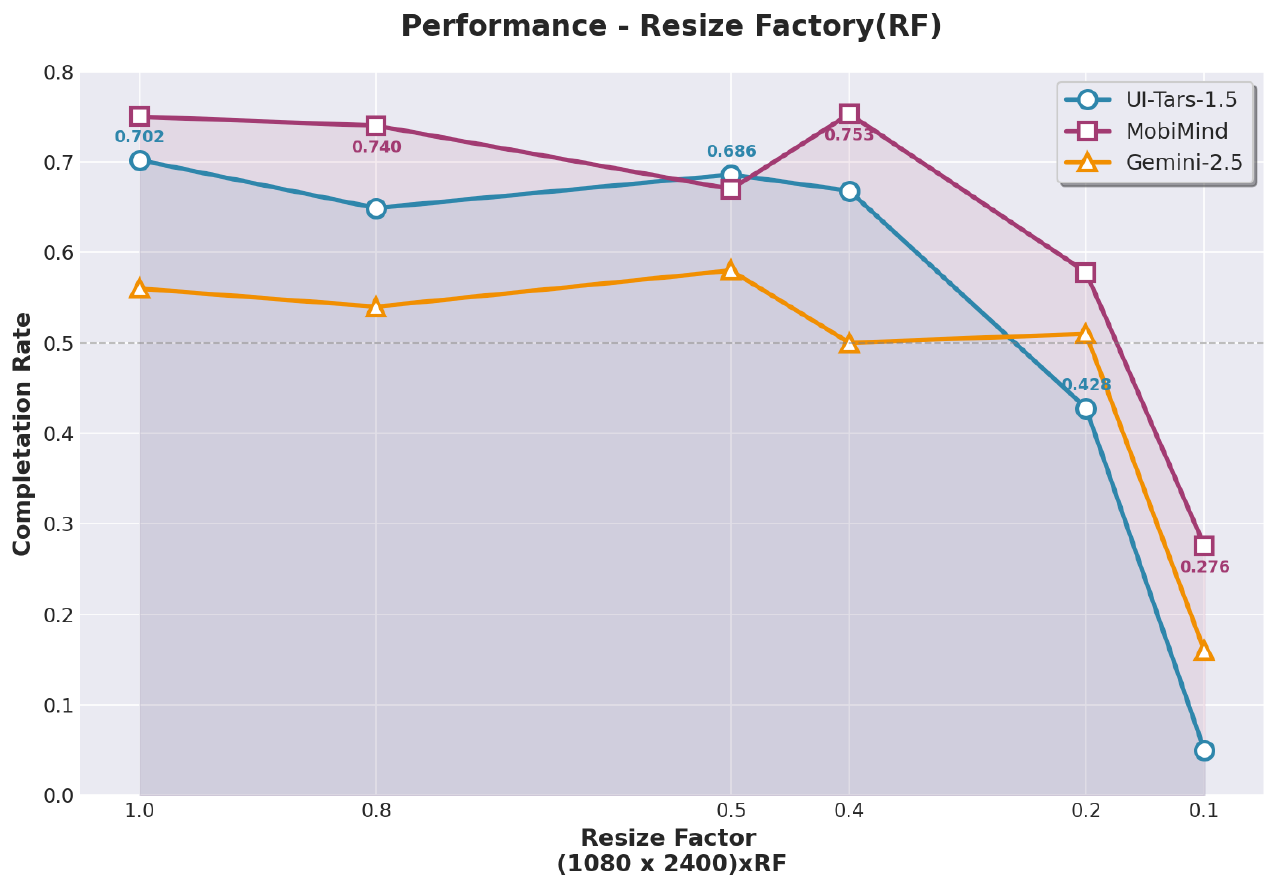}}
    \caption{
      \textbf{Completion rate of different models under varying resolution scaling factors}. We randomly sample representative examples and evaluate them at different scaling levels based on a resolution of 1080×2400.
    }
    \label{fig:rf}
\end{figure}
We evaluate UI-TARS-1.5, MobiMind, and Gemini-2.5-Flash on the same evaluation subset by scaling images with an original resolution of 1080×2400 to different factors and assessing task completion performance. We observe that when the scaling factor falls below 0.2, all models exhibit a pronounced drop in completion rate.

\end{document}